\documentclass[dvipsnames]{article}

\usepackage[final,nonatbib]{sty/neurips2024/neurips_2024}

\usepackage[utf8]{inputenc} % allow utf-8 input
\usepackage[T1]{fontenc}    % use 8-bit T1 fonts
\usepackage{hyperref}       % hyperlinks
\usepackage{url}            % simple URL typesetting
\usepackage{booktabs}       % professional-quality tables
\usepackage{amsfonts}       % blackboard math symbols
\usepackage{nicefrac}       % compact symbols for 1/2, etc.
\usepackage{microtype}      % microtypography
\usepackage{xcolor}         % colors

\usepackage{subfigure} % subfigure
\usepackage{multirow}
\usepackage{makecell}
\usepackage{algorithm}
\usepackage[noend]{algpseudocode}
\usepackage{seqsplit}
\usepackage{minibox}
\usepackage{booktabs}       % professional-quality tables
\usepackage{amsfonts}       % blackboard math symbols
\usepackage{nicefrac}       % compact symbols for 1/2, etc.
\usepackage{amssymb}
\usepackage{enumitem} % leftmargin
\usepackage{wrapfig}

\usepackage{tikz}
\usetikzlibrary{shapes,backgrounds,arrows,fit,calc,patterns}

\newcommand{\para}[1]{\textbf{#1}}
\newcommand{\BMK}[1]{\textcolor{red}{[BOOKMARK]}}

\newcommand{\sgsup}{\texttt{SGen}$^\texttt{Sup}$}
\newcommand{\sgsups}{\sgsup~}
\newcommand{\sgsemi}{\texttt{SGen}$^\texttt{Semi}$}
\newcommand{\sgsemis}{\sgsemi~}
\newcommand{\sgem}{\texttt{SGen}$_\texttt{EM}$}
\newcommand{\sgems}{\sgem~}
\newcommand{\sgsemisup}{\texttt{SGen}$^{\texttt{Semi-Sup}}_\texttt{NoMS}$}

\newcommand{\sgpl}{\texttt{SGen}$_\texttt{PL}^\texttt{H-Semi}$}
\newcommand{\sgpls}{\sgpl~}
\newcommand{\sgpfl}{\texttt{SGen}$_\texttt{PFL}^\texttt{H-Semi}$}
\newcommand{\sgpfls}{\sgpfl~}
\newcommand{\sgseminoms}{\texttt{SGen}$^{\texttt{Semi}}_\texttt{NoMS}$}
\newcommand{\sgseminomss}{\sgseminoms~}

\newcommand{\ie}{\emph{i.e.,}~}
\newcommand{\eg}{\emph{e.g.,}~}

\usepackage{pifont}
\providecommand{\para}[1]{\vspace{2ex}\noindent\textbf{#1}}

\usepackage{amsmath} % http://ctan.org/pkg/amsmath
\usepackage{bm} % for bold symbols like \bm{\alpha} or {\bm \alpha\beta}
\usepackage{bbm} % for mathbbm
\usepackage{amsthm}
\usepackage{mathtools}

\newcommand{\ep}{\varepsilon}
\renewcommand{\epsilon}{\ep}

\makeatletter
\@ifundefined{proposition}{%
    
}{}
\@ifundefined{definition}{%
}{}
\@ifundefined{lemma}{%
    \newtheorem{lemma}{Lemma}
}{}
\@ifundefined{corollary}{%
    \newtheorem{corollary}{Corollary}
}{}
\@ifundefined{theorem}{%
    \newtheorem{theorem}{Theorem}
}{}
\@ifundefined{corollary}{%
    
}{}
\@ifundefined{assumption}{%
    
}{}
\@ifundefined{problem}{%
    
}{}
\@ifundefined{problem*}{%
    \newtheorem*{problem*}{Problem}
}{}
\@ifundefined{claim}{%
    
}{}
\@ifundefined{example}{%
    
}{}
\@ifundefined{implication}{%
    
}{}
\@ifundefined{remark*}{%
    \newtheorem*{remark*}{Remark}
}{}
\makeatother

\newtheorem{innercustomgeneric}{\customgenericname}
\providecommand{\customgenericname}{}
\newcommand{\newcustomtheorem}[2]{%
  \newenvironment{#1}[1]
  {%
   \renewcommand\customgenericname{#2}%
   \renewcommand\theinnercustomgeneric{##1}%
   \innercustomgeneric
  }
  {\endinnercustomgeneric}
}
\newcustomtheorem{customclaim}{Claim}

\providecommand{\realnum}					{\mathbb{R}}
\providecommand{\integernum}				{\mathbb{Z}}

\renewcommand{\(}						{\left(}
\renewcommand{\)}						{\right)}

\providecommand{\Prob}{\mathbbm{P}}

\providecommand{\Exp}{\mathbbm{E}}

\def\b{\mathbf{b}}

\def\h{\mathbf{h}}

\def\p{\mathbf{p}}

\def\w{\mathbf{w}}
\def\x{\mathbf{x}}
\def\y{\mathbf{y}}

\def\I{\mathbf{I}}

\def\W{\mathbf{W}}
\def\X{\mathbf{X}}

\def\Z{\mathbf{Z}}

\def\eh{\hat{{e}}}

\def\kh{\hat{{k}}}
\def\lh{\hat{{l}}}

\def\sh{\hat{{s}}}

\def\yh{\hat{{y}}}

\def\Ch{\hat{{C}}}

\def\Eh{\hat{{E}}}

\def\Sh{\hat{{S}}}

\def\Uh{\hat{{U}}}

\def\As{\mathcal{{A}}}

\def\Ds{\mathcal{{D}}}
\def\Es{\mathcal{{E}}}

\def\Hs{\mathcal{{H}}}

\def\Ms{\mathcal{{M}}}

\def\Qs{\mathcal{{Q}}}
\def\Rs{\mathcal{{R}}}

\def\Vs{\mathcal{{V}}}
\def\Ws{\mathcal{{W}}}
\def\Xs{\mathcal{{X}}}
\def\Ys{\mathcal{{Y}}}

\title{Selective Generation for \\ Controllable Language Models}

\author{%
  Minjae Lee$^*$
  \\
  GSAI
  \\
  POSTECH
  \\
  \texttt{minjae.lee@postech.ac.kr} \\
  \And
  Kyungmin Kim$^*$
  \\
  GSAI
  \\
  POSTECH
  \\
  \texttt{kkm959595@postech.ac.kr}
  \AND
  Taesoo Kim
  \\
  SCS \& SCP
  \\
  GaTech
  \\
  \texttt{taesoo@gatech.edu}
  \And
  Sangdon Park
  \\
  GSAI \& CSE
  \\
  POSTECH
  \\
  \texttt{sangdon@postech.ac.kr} 
}
\begin{document}

\maketitle
\begin{NoHyper}
\def\thefootnote{*}\footnotetext{Equal contribution}
\end{NoHyper}

\begin{abstract}

Trustworthiness of generative language models (GLMs) is crucial in their deployment to critical decision making systems. 
Hence, certified risk control methods such as selective prediction and conformal prediction have been applied to mitigating the hallucination problem in various supervised downstream tasks. 
However, the lack of appropriate correctness metric hinders {applying} such principled methods to language generation tasks. 
In this paper, we circumvent this problem by leveraging the concept of \textit{textual entailment} to evaluate the correctness of the generated sequence, and propose two selective generation algorithms which control the false discovery rate with respect to the textual entailment relation (FDR-E) with a theoretical guarantee: \sgsups and \sgsemi. 
\sgsup, a direct modification of the selective prediction, is a supervised learning algorithm which exploits entailment-labeled data, annotated by humans. 
Since human annotation is costly, we further propose a semi-supervised version, \sgsemi, which fully utilizes the unlabeled data by pseudo-labeling, leveraging an \textit{entailment set function} learned via conformal prediction.
Furthermore, \sgsemis enables to use more general class of selection functions, \textit{neuro-selection functions}, and provides users with an optimal selection function {class} given multiple candidates. 
Finally, we demonstrate the efficacy of the \texttt{SGen} family in achieving a desired FDR-E level with comparable selection efficiency to those from baselines on both open and closed source GLMs.
Code and datasets are provided at \url{https://github.com/ml-postech/selective-generation}.

\end{abstract}

\vspace{-1ex}
\section{Introduction}
\vspace{-1ex}

% pros: GLMs for human-level language generation
Generative language models (GLMs) \cite{radford2019language,brown2020language,touvron2023llama,alpaca} have garnered significant attention for their ability to generate human-level language \cite{chatgpt}
primarily due to underlying transformer architectures \cite{vaswani2017attention}.
% cons: hallucinations in GLMs
However, GLMs raise concerns about generating hallucinated facts \cite{li2020dont},
which is an undesirable property when they are used as knowledge retrieval sources.
% solution 1: GLM fine-tuning (w/ model retraining)
{This issue can be mitigated by fine-tuning  with
human feedback \cite{li2020dont,ouyang2022training}, but it remains expensive in terms of training and labeling costs.}
% solution 2: certified risk control methods (w/o model retraining)
Certified risk control methods such as selective prediction \cite{geifman2017selective} and conformal prediction \cite{vovk2005algorithmic} are promising cost-efficient alternatives, which have been applied to the hallucination mitigation in various supervised downstream tasks \cite{geifman2017selective,vovk2005algorithmic,Park2020PAC,bates2021distribution,gibbs2021adaptive,park2023acon2}.

% metirc misalignment is the main bottleneck applying certified methods to language generation
The main bottleneck in applying such certified methods to language generation tasks is that provided risk control guarantees require correctness labels during the learning process.
% why metric misalignment occurs in language generation
Specifically, in classification, {high-quality correctness labels can be directly acquired} by comparing true and predicted labels using exact match (EM). However, this is not the case for language generation tasks, since multiple valid answers can exist for the same question.
% need for the evaluation metric taking multiplicity of valid sequences in language generation into consideration
As correctness metrics such as EM and F1-score do not account {for} the multiple valid answers, directly applying them to language generation tasks results in a significant gap between the true and measured correctness, which we call the \textit{metric misalignment}. Thus, a correctness evaluation metric that accounts for multiple answers is required.

\begin{figure}
  \centering
  \includegraphics[width=0.8\linewidth]{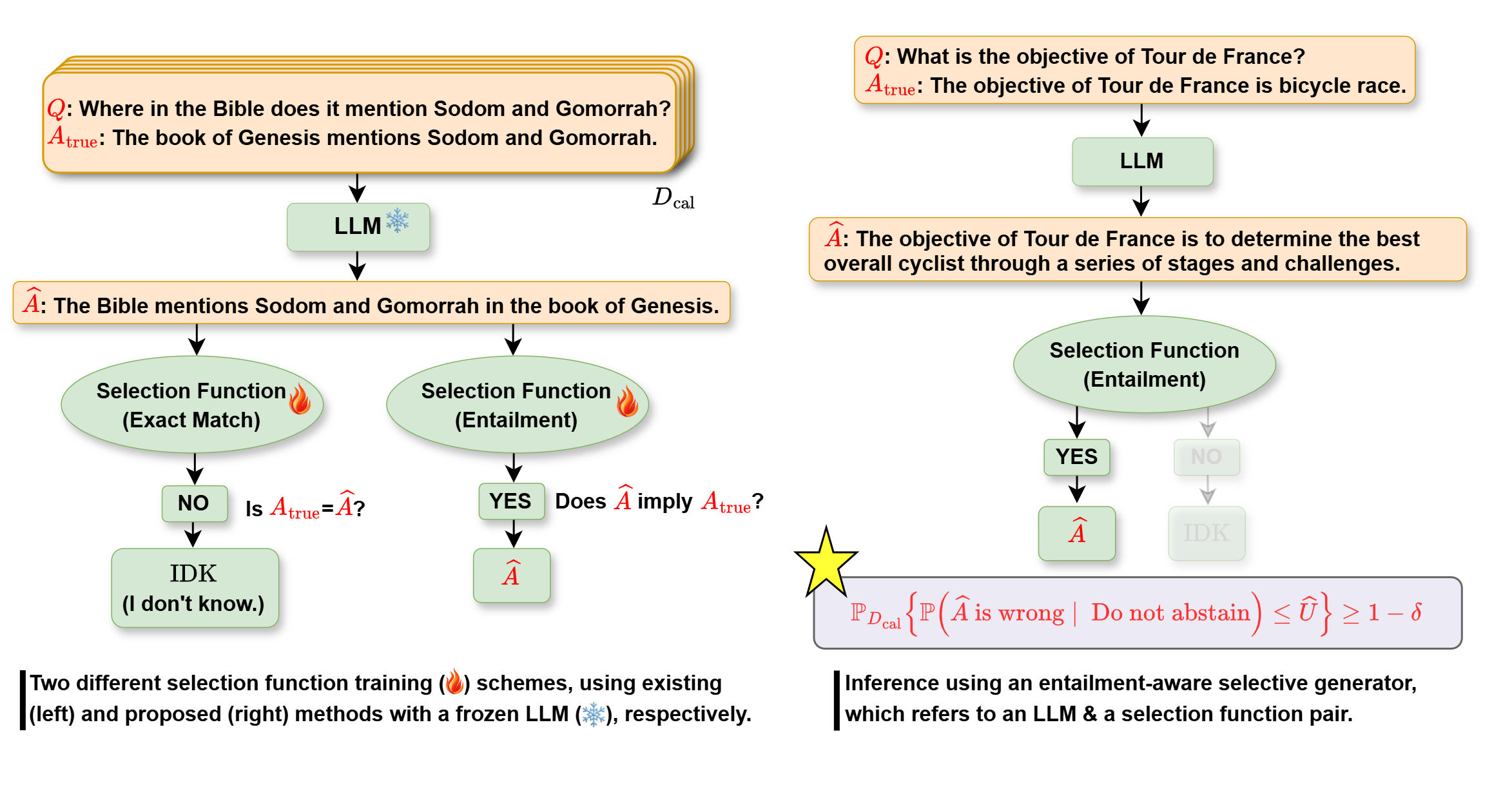}
  \caption{
    An overview and qualitative results of our  method with GPT-3.5-Turbo. The crux is to learn an entailment-aware selective generator with an abstaining option that controls the rate of hallucination (in a false discovery rate) over generated sequences with a probabilistic guarantee.
    % \SP{need larger font size. remove space between lines}
    % \textcolor{purple}{[KK: Modified. Please check and provide me with a feedback.]}
    % {\color{red} [MJ: How about .pdf?]}
    % \SP{looks comments are addressed}
    }
  \label{fig:method_overview}
  \vspace{-4ex}
\end{figure}

% circumvent metric misalignmen issue via textual entailment relation
In this paper, we resolve the metric misalignment problem by leveraging \textit{textual entailment} to evaluate the correctness of generated answers and define the false discovery rate with respect to the textual entailment relation (FDR-E). 
Given two ordered sequences, a premise and a hypothesis, we say that the premise entails the hypothesis if the hypothesis is true given the premise. 
% propose two selective generator learning algorithms based on textual entailment
Based on this {notion of} entailment, we propose two selective generation algorithms, \sgsups and \sgsemi, which are generalized versions of selective classification \cite{geifman2017selective} to control the FDR-E by abstaining from returning an answer when a GLM is uncertain of its answer.

% supervised selective generator learning algorithm
In particular, \sgsup, a direct modification of \cite{geifman2017selective}, is a supervised selective generator learning algorithm which requires entailment labels. 
This necessitates human annotations on textual entailment, where a generated answer is the premise and a true answer is the hypothesis. 
% semi-supervised selective generator learning algorithm
As labeling is expensive and \sgsups solely relies on entailment-labeled data, we propose a semi-supervised method, \sgsemi, which enables the exploitation of entailment-unlabeled data in learning a selective generator by pseudo-labeling textual entailment using an \textit{entailment set function} learned via conformal prediction \cite{vovk2005algorithmic}. 
% leveraging entailment classifier for CP-based pseudo-labeling 
Based on an entailment classifier originally developed for the natural language inference problem \cite{bowman2015large, williams2018broad}, the estimated entailment set function approximates a true entailment set function, which returns all entailed answers if a true answer is given as a hypothesis.

% from single-threshold indicator ftn. with the given confidence-rate ftn. to the multiple-threshold one with learned features 
Additionally, \sgsemi ~introduces the general class of selection functions for selective generation, called \textit{neuro-selection functions}. In selective prediction, learning a selective predictor is equivalent to learning a selection function, which is an indicator function to decide whether to abstain from returning a prediction. The standard selective prediction algorithm \cite{geifman2017selective} considers the class of single-threshold indicator functions using a pre-specified confidence-rate function. 
% selection efficiency & confidence-rate ftn. quality
For the same risk level, the better the confidence-rate function quantifies the model's uncertainty, the less likely the selective predictor is to abstain from making a prediction. 
We refer to this as \textit{selection efficiency} henceforth. 
As appropriate confidence calibration for language generation remains challenging, optimizing a single-threshold indicator function with a poorly calibrated confidence-rate function leads to low selection efficiency. 
Instead, we generalize the selection function by using a multiple-threshold indicator function with trainable features.
% provide model selection to users
{Furthermore, \sgsemis provides a user with an optimal class of selection functions among possible candidates in terms of the FDR-E.}
% existing selection ftns: single-threshold indicator ftn. with the given confidence-rate ftn.
% In selective prediction, learning a selective predictor is equivalent to learning a selection function, which is an indicator function to decide whether to abstain from returning a prediction. The standard selective prediction algorithm \cite{geifman2017selective} and similar algorithms \cite{vovk2005algorithmic,papadopoulos2002inductive,Park2020PAC,angelopoulos2021uncertainty,bates2021distribution,park2023acon2} consider the class of single-threshold indicator functions using a pre-specified confidence-rate function. 
% % selection efficiency & confidence-rate ftn. quality
% For the same risk level, the better the confidence-rate function quantifies the model's uncertainty, the less likely the selective predictor is to abstain from a prediction. 
% We refer to this as \textit{selection efficiency} henceforth.

% poorly calibrated confidence-rate ftn. in language generation => generalize selection ftn

Finally, we empirically demonstrate the efficacy of \sgsemis over open and closed source GLMs, where
we consider \sgsups as one of our baselines as it is a direct modification of \cite{geifman2017selective}.
To validate our method and its theoretical guarantee, we create a new dataset on textual entailment using the Natural Questions (NQ) dataset \cite{kwiatkowski2019natural} for each GLM. {Given a question and answer pair, the textual entailment is labeled by letting a generated answer as a premise and the true answer in declarative form as a hypothesis}. 
% In particular, we annotate textual entailment on each question and answer pair in the dataset for each GLM. 
{As communities lack human-annotated entailment-labeled data for language generation}, we believe that our dataset contributes to the hallucination evaluation of GLMs. 
% based on the transformation method by \cite{demszky2018transforming}, which converts the question and answer pair in a question-answering dataset into a declarative form, textual entailment is annotated by using a generated answer as the premise and a true answer in a declarative form as the hypothesis. 
For both open and closed source GLMs, \sgsemis is effective in achieving a desired FDR-E level with better selection efficiency compared to baselines.

\vspace{-1ex}
\subsection{Related Work}
\vspace{-1ex}

{We introduce two main research directions to mitigate hallucination in GLMs.} 

\para{Heuristics for hallucination mitigation.} The hallucination in language generation usually refers to the situation where a GLM generates wrong answers with high confidence, which hinders the reliable deployment of GLMs. As fine-tuning methods are expensive, heuristics for hallucination mitigation without tuning have been proposed \cite{jiang2021howcan,manakul2023selfcheckgpt}. 
Notably, \cite{manakul2023selfcheckgpt} proposes a performant hallucination detection method, which quantifies the self-consistency among multiple generated answers for the same question using textual entailment models to detect the hallucination. 
However, these methods do not provide certified control over the occurrence of hallucinated contents.

\para{Certified methods for hallucination mitigation.} Conformal prediction outputs a prediction set that is guaranteed to contain a true label with high probability, where a provided coverage guarantee is model-agnostic under a mild assumption on a data \cite{vovk2005algorithmic}. 
Although this property enables the safe deployment of complex models and has made conformal prediction popular \cite{vovk2005algorithmic,bates2021distribution,gibbs2021adaptive,vovk2013conditional,fisch2021fewshot,park2022pacmeta}, the constructed prediction sets in language generation are often less-informative due to an unbounded label space, which frequently renders the coverage guarantee ineffective \cite{quach_conformal_2024,mohri2023learning}. 
To restrict the prediction set to a moderate size, \cite{quach_conformal_2024} constructs the prediction set over answers 
by sampling them sequentially, while still satisfying the coverage guarantee. 
%which is sequentially enlarged by generating multiple output sequences one-by-one until the pre-specified stopping rule is achieved. 
% Thought this avoids the unbounded label space issue, 
{Still}, post-selection of answers from the prediction set is necessary for final decision making, which may result in the selection bias \cite{jin_selection_2023, jin_confidence_2024}. 
% \cite{mohri2024language, cherian_large_2024} decompose generated answers into sub-claims via conformal prediction with alignment labels whether the sub-claims contain no contradiction.
{\cite{mohri2024language, cherian_large_2024} decompose generated answers into alignment-labeled sub-claims and return a set of sub-claims that contains no contradiction with high probability via conformal prediction.}
Even though the post-selection is unnecessary, it requires expensive alignment labels for every {sub-claim}.
%and it does not give information about whether the subset of sub-claims as a whole is appropriate or not. 

{Unlike} conformal prediction, selective prediction directly manages target risk at a desired level by introducing an abstaining option on unsure predictions. 
\cite{geifman2017selective} proposes a selective prediction method mainly for classification, which learns a threshold-based selection function that controls the false discovery rate (FDR) to a desired level. 
\cite{mohri2023learning} generalizes the selective prediction to language generation. 
However, their theoretical guarantee is not focused on the target risk to control, but on a consistency property of a surrogate loss function with respect to a true loss function in optimization process. 
\cite{gui_conformal_2024}, concurrently published {with} our paper, proposes a certified selective generation method for context-given language generation which controls the FDR. 
Unlike \cite{geifman2017selective} which takes the number of selected samples as constraint in learning the selection function, \cite{gui_conformal_2024} set the power as constraint. 
However, as \cite{mohri2023learning} does, they require an additional calibration set for training an entailment scoring function.
Importantly, while existing selective generation methods are supervised learning methods, we propose a semi-supervised learning algorithm that can fully leverage entailment-unlabeled data.

\vspace{-1ex}
\section{Background}
\vspace{-1ex}

While we consider general language generation tasks, we confine our scope to the open-ended question-answering task and define the notation accordingly for the sake of clarity and for maintaining consistency in descriptions on the experiment. 
Specifically, let $\mathcal{W}$ denote a token space constructed using a tokenizer, such as Byte Pair Encoding \cite{gage1994new}, and let $\Ws^*$ denote a token sequence space, defined as $\Ws^* \coloneqq \cup_{i=0}^\infty \Ws^i$. 
Let $( \x, \y ) \in \mathcal{X} \times \mathcal{Y}$ be a question and answer sequence pair, where $\mathcal{X} \coloneqq \Ws^*$ and $\mathcal{Y} \coloneqq \Ws^*$ refer to the token sequence spaces of questions and answers, respectively. 
We assume the answer sequence is in a declarative form.
Finally, $\x_{i:j}$ refers to the sub-sequence of $\x$ from the $i$-th to the $j$-th token.

\vspace{-1ex}
\subsection{Language Generation}
\vspace{-1ex}

Given a question as input, a GLM generates an answer through the sequential process called decoding, which we call language generation. 
Here, we consider the greedy decoding, a deterministic generation process described as follows. 
Let $p_M: \Xs \times \Ws \to \realnum_{\ge 0}$ denote a GLM which returns a next-token distribution given the input sequence $\x$, where $\sum_{w \in \Ws} p_M(w \mid \x) = 1$ for all $\x \in \mathcal{X}$. 
A language generator $G: \mathcal{X} \to \mathcal{Y}$ using greedy decoding sequentially generates tokens from the GLM as follows:
$
    \hat{\y}_i \coloneqq \arg\max_{w\in \Ws}~p_M(w \mid (\x, \hat\y_{1:i-1}))
$
for $i \ge 2$ {and}
$\hat{\y}_1 \coloneqq {\arg\max}_{w\in \Ws}~p_M(w \mid \x)$. 
The generator $G$ returns a generated answer $\hat{\y} \coloneqq G( \x )$ and terminates the decoding process when the end-of-sequence (\texttt{EOS}) token is returned. 
% \begin{align*}
%     \hat{y}_i \coloneqq \begin{cases}
%         \underset{w\in \Ws}{\arg\max}~p_M(w \mid \x) &\text{if~} i = 1 \\
%         \underset{w\in \Ws}{\arg\max}~p_M(w \mid (\x, \hat\y_{1:i-1})) & (i \geq 2),
%     \end{cases}
% \end{align*}
% returning the generated answer $\hat{\y} \coloneqq G( \x )$ and terminating the decoding process when the end-of-sequence token (\texttt{EOS}) is returned. 
Here, the conditional probability of the answer $\hat{\y}$ is defined as $f_{M}(\x, \hat\y) \coloneqq p_M(\hat\y_1 \mid \mathbf{x})\prod_{i=2}^{\vert\hat{\mathbf{y}}\vert}p_M(\hat\y_i \mid (\mathbf{x}, \mathbf{\hat{y}}_{1:i-1}))$, commonly used as its uncertainty measure.

%% \begin{itemize}
%% \item token space: $\Ws$
%% \item a sequence of tokens: $\Xs \coloneqq \Ws^*$
%% \item language model $p: \Xs \times \Ws \to \realnum_{\ge 0}$: a distribution over tokens given a token sequence
%% \item language generator $G: \Xs \to \Xs$
%% \item the probability of a generated sequence: $\prod_{i=1}^{|\y|} p(\y_i \mid \x)$, where $\y = G(\x)$
%% \item we consider greedy decoding
%% \end{itemize}

%
%
\vspace{-1ex}
\subsection{Selective Prediction}
\vspace{-1ex}

Selective prediction refuses to make a prediction by returning ``I don't know'' (\texttt{IDK}) if the prediction is uncertain. 
In classification, the selective classifier $\hat{S}$ consists of a pair of a classifier $\yh$ and a selection function $\sh$, and is defined as follows:
$
    \hat{S}(\x) \coloneqq \begin{cases}
        G(\x) &\text{if~} \hat{s}(\x) = 1 \\
        \texttt{IDK} & \text{otherwise}
    \end{cases},
$
% $\hat{S}(\x)$ returns  $G(\x) \text{~if~} \hat{s}(\x) = 1$ or returns $\texttt{IDK}$ otherwise,
where 
%$\hat{y}(\x) \coloneqq \underset{y \in \mathcal{Y}}{\arg\max}~f(\x, y)$. 
$\hat{y}(\x) \coloneqq {\arg\max}_{y \in \mathcal{Y}}~f(\x, y)$. 
Here, $f(\x, y)$ refers to an estimated likelihood of the given input $\x$ for being a class $y$, determined by an underlying classification model $f$. 
Although the selection function can be of arbitrary form, the common choice is a single threshold indicator function using the maximum likelihood as the confidence-rate function, \ie $\hat{s}(\x) \coloneqq \mathbbm{1}( f(\x, \hat{y}) \geq \tau )$. 
Here, the confidence-rate function is defined to quantify the uncertainty of the model's prediction. 
Under the independent and identically distributed (i.i.d.) assumption,
\cite{geifman2017selective} proposed the certified threshold learning algorithm which controls the false discovery rate (FDR) with respect to the EM metric with the PAC guarantee, where the FDR is defined as
$
    \mathcal{R}_{\text{EM}}(\hat{S}) \coloneqq \mathbbm{P}\{ \hat{y}(\x) \neq y \mid \hat{S}(\x) \neq \texttt{IDK} \}.
$
Since EM considers the answer $\hat{y}(\x)$ to be correct when it is exactly the same as the reference answer $y$, it is an inappropriate correctness metric for language generation problems that can have multiple valid sequences for the same input. 
This results in learning a too conservative and vacuous selection function for language generation, which is empirically verified by our experiments. 
Thus, we leverage the textual entailment to evaluate the correctness of the generated sequence to alleviate the metric misalignment problem.
%, which requires costly human annotations on textual entailment.

\vspace{-1ex}
\subsection{Textual Entailment}
\vspace{-1ex}
Natural language inference (NLI), also denoted as recognizing textual entailment, predicts whether one sequence implies another. 
The former refers to a premise ($\p$), and the latter refers to a hypothesis ($\h$). 
Since the release of two large-scale benchmarks of ordered sequence pairs labeled with textual entailment \cite{bowman2015large, williams2018broad}, a number of transformer-based entailment classifiers have been proposed and shown impressive results. 
Each pair is classified into one of three categories: \textit{entailment} if $\h$ is true given $\p$; \textit{contradiction} if $\h$ is false given $\p$; and \textit{neutral} otherwise. 
In this paper, we define the entailment scoring function as $f_E( G(\x), \y ) \coloneqq 1 - p_E( \text{\textit{contradict}} ~\vert~ \p=G(\x), \h=\y )$ to estimate and pseudo-label the correctness of $G(\x)$, where $p_E( \text{\textit{contradict}} ~\vert~ \p=G(\x), \h=\y )$ is the likelihood that $G(\x)~\text{contradicts}~\y.$ 
While pseudo-labeling enables the full exploitation of unlabeled data to learn a selection function, controlling the mislabeling error remains as a challenge.

\vspace{-1ex}
\subsection{Conformal Prediction}
\label{sec:pacps}
\vspace{-1ex}
Conformal prediction \cite{vovk2005algorithmic} outputs a prediction set to quantify the uncertainty of a given model with a model-agnostic correctness guarantee under minimal assumptions on data generating process.
Specifically, under the i.i.d. assumption, PAC conformal prediction \cite{Park2020PAC} incorporates the interpretation of tolerance regions \cite{wilks1941determination} and training-conditional inductive conformal prediction \cite{vovk2013conditional} through the lens of PAC learning theory \cite{valiant1984theory}.
In this paper, we adopt the PAC prediction set learning algorithm to control the rate of mislabeling error in pseudo-labeled samples used to learn a selection function for selective generation. See \autoref{apdx:disc:cp} for detailed discussion on conformal prediction.

% conformal set model
\para{Scalar-parameterized Conformal Set.}
In this paper, we consider a conformal set $C: \Xs \to 2^{\Ys}$ parameterized by a scalar \cite{Park2020PAC, papadopoulos2002inductive} as
$
%\begin{equation}
  C(\x) \coloneqq \left\{ y \in \Ys \mid f(\x, y) \ge \tau \right\},
  %\label{eq:ps}
%\end{equation}
$
where $\tau \in \Hs$ is a scalar parameter to learn, $\Hs$ is a hypothesis space (\eg $\Hs$ a {finely} discretized non-negative real numbers), and $f: \Xs \times \Ys \to \realnum_{\ge 0}$ is called a \textit{scoring function}. 
The scoring function corresponds to a target model whose uncertainty is to be quantified, where the softmax output is a common choice in classification. 
Specifically, $f(\x, y)$ measures the {likelihood} of $y$ as a response given $\x$ as input. 
% \eg the softmax output in classification.
% In relation to the size of conformal sets, 
% it 
% will be relatively large if the target model $f$ is uncertain about the prediction of the given input, \ie small $f(\x, y)$, containing multiple labels whose softmax probabilities exceed the threshold.

%Taking the classification problem as an example, the size of the conformal set will be relatively large if the target model is uncertain about the prediction of the given input, containing multiple labels whose softmax probabilities exceed the threshold.

%% error
\para{PAC Guarantee.}
The PAC prediction set learning algorithm outputs a conformal set $\hat{C}$ which upper bounds a miscoverage rate 
$
  %\label{eq:miscoverage}
  \Rs_{\text{MC}}(\Ch) \coloneqq \Prob\{ y \notin \Ch(\x) \}
$
to a desired level $\epsilon \in (0, 1)$, where the miscoverage rate can be generalized to risk
$
  %\label{eq:miscoverage}
  \Rs_{\text{01}}(\Ch) \coloneqq \mathbbm{E}\{ \ell_{01}( \hat{C}, \x, y ) \}, 
$
on any indicator losses that are monotonic with respect to $\tau$. 
The algorithm is \textit{probably approximately correct} (PAC) in the sense that it provides a calibration data-conditional guarantee at every risk and confidence level. 
Specifically, it controls the risk to a desired level irrespective of which calibration data is used to learn $\hat{C}$ with a desired confidence $\delta \in (0, 1)$ as follows:
$
  \Prob\{ \Rs_{01}(\Ch) \le \epsilon \} \ge 1 - \delta,
$
where the probability is taken over the calibration set $\Z \sim \mathcal{D}^{n}$ to learn the conformal set. 
In this paper, we leverage the PAC conformal set for a pseudo-labeling function such that 
the guarantee on the labeling quality provides the overall PAC guarantee in semi-supervised selective generator learning algorithm. 

%In this paper, to pseudo-label textual entailment, we learn the PAC conformal set by defining the indicator loss to be 1 if the generated sequence is contradictory to a given question
%and by letting the entailment scoring function be the target model.

%% algorithm
\para{Algorithm.}
The PAC conformal set learning algorithm  
$\mathcal{A}_{\text{Binom}}: (\Xs \times \Ys)^* \to \Hs$ \cite{Park2020PAC,vovk2013conditional,Park2022PAC} returns the conformal set parameter $\hat{\tau}$, where $\Hs$ is a {finely}-discretized $\realnum_{\ge 0}$.
Specifically, the algorithm returns $\hat{\tau}= \max_{\tau \in \Hs}~\tau$ subject to $U_{\text{Binom}}(k_\tau; n, \delta) \le \epsilon$, where 
$k_\tau \coloneqq \sum_{i=1}^n \ell_{01}(\Ch, \x_i, y_i)$. 
% is mistakes by the conformal set $\Ch$.
Letting $F(k; n, \theta)$ be a cumulative distribution function of a binomial distribution with $n$ trials and success probability $\theta$,
$U_{\text{Binom}}(k; n, \delta) \coloneqq \inf \left\{ \theta \in [0, 1] \mid F(k; n, \theta) \le \delta \right\} \cup \{1\}$ is an upper binomial tail bound that satisfies
$
  \Prob\{ \Rs_{01}(\Ch) \le U_{\text{Binom}}(k_\tau; n, \delta) \} \ge 1 - \delta,
$
where $\delta$ is the desired confidence.
Note that we similarly denote a lower binomial tail bound by $L_\text{Binom}$.
If optimization in the algorithm $\As_\text{Binom}$ is infeasible, the algorithm returns $\hat{\tau}=0$, a vacuous conformal set. 
Thus, the algorithm is PAC, and see \autoref{apdx:disc:cp} for proof.

% Here, $U_{\text{Binom}}(k_\tau; n, \delta)$ satisfies 
% \begin{align*}
%   \Prob\{ \Rs_{01}(C) \le U_{\text{Binom}}(k_\tau; n, \delta) \} \ge 1 - \delta,
% \end{align*}
% where $U_{\text{Binom}}(k; n, \delta) \coloneqq \inf \left\{ \theta \in [0, 1] \mid F(k; n, \theta) \le \delta \right\} \cup \{1\}$ and $F(k; n, \theta)$ is a cumulative distribution function of a binomial distribution with $n$ trials and success probability $\theta$. 
% If the optimization is infeasible, the algorithm returns $\hat{\tau}=0$, a vacuous conformal set. 
% The algorithm is PAC, and see \autoref{apdx:disc:cp} for proof.
%
%
%
\vspace{-1ex}
\subsection{Calibration}
\vspace{-1ex}
In classification, calibration aims to adjust the classifier's maximum likelihood response, or confidence, to be correct. 
We say the classifier response $f: \mathcal{X} \times \mathcal{Y} \to \mathbb{R}_{\ge 0}$ is \textit{perfectly calibrated} with respect to a distribution $\mathcal{D}$ over $\mathcal{X} \times \mathcal{Y}$ and a classifier $\hat{y}$ if
$
  \Prob\left\{ \y = \yh(\x) \mid f(\x, \yh(\x)) = t \right\} = t 
$
for all $t \in [0, 1]$ \cite{degroot1983comparison,zadrozny2002transforming}. 
Calibration aims to find the classifier response such that it is perfectly calibrated asymptotically. 
In this paper, we make an interesting connection between calibration and selective {generation}. In particular, given the definition of the perfect calibration for a language scoring function $f_M$, we formally provide a sufficient condition for a selective generator to control the FDR with respect to the textual entailment relation at \emph{any} desired risk level.
%Taking it a step further, we generalize the class of selection functions from the signle-threshold indicator function using the calibrated scoring function to the multiple-threshold indicator function using learnable features.
\vspace{-1ex}
\section{Problem: Selective Generation}
\vspace{-1ex}

% definitions
% Let $\Ws$ be a token space.

% % main goal
% Our main goal is to find a learning algorithm that returns a selective generator given a calibration set.
% and
% a distribution over labeled examples $\Xs \times \Ys$ by $\Ds$

% In particular,
% given 
% a generator $G: \Xs \to \Ys$,
% we consider a \emph{selective generator} $\smash{\Sh}: \Xs \to \Ys \cup \{\texttt{IDK}\}$,
% which abstains from answering (denoted by $\texttt{IDK}$) if the generated answer $G(\x)$ is unsure with respect to a selection function $\sh(\x, G(\x)) \in \{0, 1\}$, \ie
% \begin{equation*}
%   \Sh(\x) \coloneqq
%   \begin{cases}
%     G(\x) & \text{if $\sh(\x, G(\x)) = 1$} \\
%     \texttt{IDK} & \text{otherwise}
%   \end{cases}.
% \end{equation*}
% In this paper, we learn a selective generator $\smash{\Sh}$ to control risk $\Rs$ defined in a generalized false discovery rate (FDR), \ie the ``inverse'' of precision,  with respect to a relation $R$, \ie
% \begin{equation}
%   \label{eq:precrisk}
%   \Rs_R(\Sh) \coloneqq 
%   \Prob\left\{ (G(\x), \y) \notin R  \;\middle|\; \Sh(\x) \neq \texttt{IDK} \right\}.
% \end{equation}

Let
$\x \in \Xs$ be a question and
$\y \in \Ys$ be an answer, 
assuming that each question has a desired answer.
Here, 
we assume $(\x, \y) {\overset{\text{i.i.d.}}{\sim}} \Ds'$, where $\Ds'$ is a data generating process of question-answering pairs. 
Then, 
given a generator $G: \Xs \to \Ys$,
we consider a \emph{selective generator} $\smash{\Sh}: \Xs \to \Ys \cup \{\texttt{IDK}\}$
which refuses to return $G(\x)$ if a selection function $\sh(\x, G(\x)) \in \{0, 1\}$ deems uncertain as follows:
\begin{equation*}
  \Sh(\x) \coloneqq
  \begin{cases}
    G(\x) & \text{if $\sh(\x, G(\x)) = 1$} \\
    \texttt{IDK} & \text{otherwise}.
  \end{cases}.
\end{equation*}
Our main goal is to learn a selective generator $\smash{\Sh}$ to control a generalized false discovery rate (FDR) with respect to a relation $R$ as
\begin{equation}
  \label{eq:precrisk}
  \Rs_R(\Sh) \coloneqq 
  \Prob\left\{ (G(\x), \y) \notin R  \;\middle|\; \Sh(\x) \neq \texttt{IDK} \right\}.
\end{equation}
 
Here, the probability is taken over examples $(\x, \y, e, v)$, where $e \coloneqq \mathbbm{1}({(G(\x), \y) \in R})$ is an additional label to be annotated due to unknown $R$ and $v \in \{0, 1\}$ is a visibility flag of $e$ for semi-supervised learning. 
For the data generation of $(\x, \y, e, v)$, 
we assume that 
a label $e$ is observed with an unknown success probability of $p_v$, independent of the generative process of $(\x, \y, e)$, \ie
$
    (\x, \y, e, v) \sim \Ds \coloneqq \Ds' \cdot \Vs,
$
where $\Ds'$ is a distribution over $\Xs \times \Ys \times \{0, 1\}$ and $\Vs \coloneqq \text{Bernoulli}( p_v )$.
Note that the definition of $e$, $\Ds'$ varies by generator $G$ even with the same data generating distribution of $(\x, \y)$.
% where the probability is taken over a labeled example $(\x, \y) \sim \Ds$ possibly along with the randomness of the generator $G$.
% Note that $\Rs_R(\Sh) = 0$ if $\Prob\{ \Sh(\x) \neq \texttt{IDK} \}=0$.
In this paper, we design a learning algorithm $\As$ that returns a selective generator $\Sh$ to control the generalized FDR with respect to $R$ within a desired level $\ep \in (0, 1)$ with probability at least $1 - \delta \in (0, 1)$, \ie
%\begin{equation}\label{eq:pacprecision}
$
  \Prob\left\{ \Rs_R(\As(\Z)) \le \ep \right\} \ge 1 - \delta.
$
%\end{equation}
Here,
the probability is taken over a calibration set $\Z \sim \Ds^n$.
This guarantee is called a probably approximately correct (PAC) guarantee \cite{valiant1984theory}.
Among selective generators that satisfies the PAC guarantee,
we choose one 
that minimizes the ratio of $\texttt{IDK}$-answers with the highest \emph{selection efficiency}.
% challenge
The main challenge is to find a sample and selection efficient PAC algorithm for any $\ep$ and $\delta$ along with designing a relation $R$ for structured labels, as in question-answering.
Frequently, we may not obtain a PAC algorithm for any $\ep$, so in this paper, we use a relaxed notion of \emph{controllable} instead of \emph{correct} if the algorithm provides minimum achievable risk beoyond a given $\ep$.

\vspace{-1ex}
\section{Semi-Supervised Learning for Controllable Selective-Generation}
\label{sec:method}
\vspace{-1ex}

In this paper, we leverage the textual entailment as the evaluation metric in language generation to consider multiple valid answers in a principled way, and propose two selective generator learning algorithms which control FDR with respect to the textual entailment: \sgsups and \sgsemi.

\vspace{-1ex}
\subsection{False Discovery Rate via Textual Entailment (FDR-E)}
\vspace{-1ex}
A textual entailment relation $R_E$ is an ordered subset of $\mathcal{Y} \times \mathcal{Y}$ where $(\y', \y) \in R_E$ if $\y'$ entails $\y$. 
In question-answering as an example, the generated answer $G(\x)$ is correct if the reference answer $\y$ is a logical consequence of $G(\x)$. In other words, $G(\x)$ is valid if $G(\x) \in E_{\text{true}}(\y)$, where the true entailment set function $E_{\text{true}}: \mathcal{Y} \to 2^{\mathcal{Y}}$ is defined as follows:
%\begin{align*}  
$
    E_{\text{true}}(\y) \coloneqq \{ \y' \in \mathcal{Y} \mid (\y', \y) \in R_E \}.
$
%\end{align*}
Then, an FDR with respect to the entailment relation $R_E$ (FDR-E) that we aim to control is as follows:
\begin{align*}  
    \mathcal{R}_{R_E}( \hat{S} ) 
    \coloneqq 
    \mathbbm{P}\{ G(\x) \notin E_{\text{true}}(\y)  \mid \hat{S}(\x) \neq \texttt{IDK} \},
\end{align*}
where the probability is taken over labeled examples, \ie $(\x, \y, e) \sim \Ds$.
Here, the label $e$ is specifically called an entailment label, \ie $e \coloneqq G(\x) \in E_\text{true}(\y)$.
% {\color{red}
% Here, the probability is taken over labeled examples $(\x, \y)$ along with additional random variables $(e, v)$. 
% In particular, we introduce $e \coloneqq G(\x) \in E_\text{true}(\y) \in \{0, 1\}$, which is an entailment label to be annotated due to unknown $E_\text{true}$ and a visibility flag $v \in \{0, 1\}$. 
% For the data generation of $(\x, \y, e, v)$, 
% we assume that 
% an entailment label $e$ is observed with probability $\mathbbm{P}\{ v=1 \}$, independent of the generative process of $(\x, \y, e) \sim \mathcal{D}$, \ie
% $
%     (\x, \y, e, v) \sim \mathcal{D} \times \text{Bernoulli}( \mathbbm{P}\{ v=1 \} ).
% $
% \SP{move to the definition.}
% }
% where the probability is taken over labeled examples, \ie $(\x, \y, e, v) \in \mathcal{X} \times \mathcal{Y} \times \mathcal{E} \times \mathcal{V}$ such that $\mathcal{E} = \mathcal{V} \coloneqq \{ 0, 1 \}$. 
% Here, we assume the data generating process that an entailment label $e$ is observed with probability $\mathbbm{P}\{ v=1 \}$, independent of the generative process of $(\x, \y, e) \sim \mathcal{D}$, \ie
% $
%     (\x, \y, e, v) \sim \mathcal{D} \times \text{Bernoulli}( \mathbbm{P}\{ v=1 \} ).
% $
% Then, for any $G$, $\mathcal{D}$, $\Vs$, and $\Sh$, the FDR-E can be decomposed as follows:
Then, for any $G$, $\mathcal{D}$, $\Vs$, and $\Sh$, the FDR-E can be decomposed as follows:
\begin{align}
\underbrace{\Prob_{\Ds_{\Sh}}\{ G(\x) \notin E_\text{true}(\y)\}}_{\text{(A)}} =
\underbrace{\Prob_{\Ds_{\Sh}}\{ v = 1 \}}_{\text{(B)}} \underbrace{\Prob_{\Ds_{\Sh}}\{ e=0 \}}_{\text{(C)}} + 
\underbrace{\Prob_{\Ds_{\Sh}}\{ v = 0 \}}_{\text{(D)}} \underbrace{\Prob_{\Ds_{\Sh}}\{ e=0 \}}_{\text{(E)}},
\label{eq:fdrdecomp-ss-ssl}
\end{align}
%where {\color{red}$\mathcal{D}_{\hat{S}} \coloneqq \Ds( \x, \y, e \mid \hat{S}(\x) \neq \texttt{IDK} )$ and $\Vs_{\Sh} \coloneqq \Vs ( v \mid \hat{S}(\x) \neq \texttt{IDK} )$}. 
where $\Prob_{\mathcal{D}_{\hat{S}}}\{ \cdot \} \coloneqq \Prob\{ \cdot \mid \hat{S}(\x) \neq \texttt{IDK})$. 
Note that as $(\x, \y, e)$ and $v$ are independent, (A), (C), and (E) in (\ref{eq:fdrdecomp-ss-ssl}) are of the same quantity, which is the target risk that we aim to find an upper bound.

\vspace{-1ex}
\subsection{FDR-E Bound for Supervised Learning}
\vspace{-1ex}

%Since entailment labels are necessary to train a certified selective generator, 
We first propose the supervised learning algorithm \sgsups (\autoref{alg:selgenel}), a direct modification of \cite{geifman2017selective} to language generation tasks. 
In particular, 
\sgsups is a supervised method in the sense that it solely exploits labeled examples $\Z_E \coloneqq \{ (\x, \y, e) \mid (\x, \y, e, v) \in \Z \wedge v=1 \}$ to learn a selective generator that controls the upper bound (C) in (\ref{eq:fdrdecomp-ss-ssl}). Note that for supervised learning, we assume that (B) in (\ref{eq:fdrdecomp-ss-ssl}) is always 1, so we only consider the the upper bound (C) via the binomial tail bound as \cite{geifman2017selective}.

\vspace{-1ex}
\subsection{FDR-E Bound for Semi-Supervised Learning}
\vspace{-1ex}

As \sgsups requires human annotations for entailment labels and makes no use of abundant unlabeled examples $\Z_U \coloneqq \{ (\x, \y) \mid (\x, \y, e, v) \in \Z \wedge v=0 \}$, we further propose a novel semi-supervised learning algorithm \sgsemis (\autoref{alg:sgsingle_additional_delta}), which fully exploits both $\Z_E$ and $\Z_U$ while controlling the FDR-E in (\ref{eq:fdrdecomp-ss-ssl}).
In particular,
we (1) estimate a true entailment set $E_{\text{true}}$ via conformal prediction with labeled examples $\Z_E$ and then (2) use the estimated entailment set $\Eh$ to annotate pseudo-labels on $\Z_U$. Finally, we (3) use both labeled and pseudo-labeled examples to learn a selective generator. 
Interestingly, this heuristic-looking algorithm could be a rigorous algorithm that controls the FDR-E of a selective generator, which will be described in the following sections.

% we exploit $\Z_E \cup \Z_U$ in learning a selective generator by estimating a pseudo-labeling function using $\Z_E$ to annotate $\Z_U$ for entailment pseudo-label. 
% The pseudo-labeling function inevitably results in mislabeling error, which challenges to form a rigorous upper bound of (E) in (\ref{eq:fdrdecomp-ss-ssl}).
% Therefore, we propose a PAC conformal set learning algorithm \texttt{ES} (\autoref{alg:es}), which returns a certified entailment set function $\hat{E}: \mathcal{Y} \to 2^{\mathcal{Y}}$ that pseudo-labels an unlabeled sample by $\hat{e} \coloneqq \mathbbm{1}( G(\x) \in \hat{E}(\y) )$ with a PAC guarantee on controlling mislabeling error. 
% \SP{connect the above summary to each subsection.}

\vspace{-1ex}
\subsubsection{FDR-E Decomposition}
\label{sec:add-decomp-semi}
\vspace{-1ex}

% a venn diagram
%\begin{figure}[t!]
\begin{wrapfigure}{r}{0.5\textwidth}
\centering
\vspace{-5ex}
\scalebox{0.6}{
\begin{tikzpicture}

% Univ
\def\Univ{(-1.5,-1) rectangle (6.5,3.7)}
% Etrue
\def\Etrue{(-1,0) rectangle (5,3)}
% Ehat
\def\Ehat{(0,-0.5) rectangle (6,2.5)}

%\draw[pattern=north east lines, fill opacity=0.5, pattern color=blue, text opacity=1] \Univ node[label={[shift={(-4,-0.2)}] above: \makecell{$\Omega$}}] {};
\draw[fill=blue!50, fill opacity=0.2, text opacity=1] \Univ node[label={[shift={(-4,-0.2)}] above: \makecell{\large $\Omega$}}] {};
\draw[pattern=north east lines, pattern color=red, fill opacity=0.5, text opacity=1] \Etrue node[label={[shift={(-5.6,-0.2)}] above: \makecell{\large $\Omega^{E_\text{true}}_\text{TD}$}}] {};
\draw[fill=gray!40, fill opacity=1, text opacity=1] \Ehat node[label={[shift={(-0.3,-0.2)}] above: \makecell{\large $\Omega_{\text{TD}}^{\Eh}$}}] {};

\begin{scope}
    \clip \Etrue;
    \fill[white] \Ehat;
\end{scope}
\draw[black] \Etrue node {};
\draw[black] \Ehat node {};
             
%\node [draw, rectangle, even odd rule, rounded corners, solid, black, minimum width=6cm, minimum height=2.5cm, fill=gray!50, align=center, label={[shift={(-0.8cm,-0.05cm)}] above left: \makecell{$E_\text{true}$}}] (Etrue) at (2,0.5) {};
% \begin{scope}
% \clip \draw[draw, rectangle, rounded corners, solid, black, minimum width=6cm, minimum height=2.5cm, fill=gray!50, align=center]  at (2,0.5) {};
% \end{scope}

% % hat{E}
% \node [draw, rectangle, rounded corners, solid, black, minimum width=6cm, minimum height=2.5cm, fill=gray!20, align=center, label={[shift={(1cm,-0.1cm)}]above right: \makecell{$\hat{E}$}}] at (3,0)   {};

% FNER
\fill (-0.5,1.7) node[below, color=red]{\textbf{FNER}};
% FER
\fill (5.5,1.2) node[below]{\textbf{FER}};
% NER
\fill (2.5,3.6) node[below, color=blue]{\textbf{NER}};

\end{tikzpicture}
}

\caption{Decomposition of a false discovery rate with respect to an entailment set $E_\text{true}$ (FDR-E).
Here, 
$\Omega_{\text{TD}}^{E}\coloneqq \{(\x, \y, e, v) \mid G(\x) \in E(\y)\}$.
}
\vspace{-2ex}
\label{fig:fdrdecomp}
%\end{figure}
\end{wrapfigure}

\sgsemis leverages unlabeled examples by estimating an entailment set as a pseudo-labeling function. However, the estimation error introduces wrong pseudo-labels. Here, we consider a rigorous way to derive the FDR-E upper bound by controlling the estimation error of the pseudo-labeling function. 
% In particular, the estimation error between the true entailment set $E_{\text{true}}$ and an estimated entailment set $\Eh$ is represented in \autoref{fig:fdrdecomp}, \ie false negative entailment rate (FNER) and false entailment rate (FER). From this, we have the following decomposition.
{In particular, two different types of estimation errors of an estimated entailment set $\Eh$ are illustrated in \autoref{fig:fdrdecomp}, \ie a false negative entailment rate (FNER) and a false entailment rate (FER). This results in the following decomposition.}
\begin{lemma}\label{lem:fdrbounddecomp}
(E) in (\ref{eq:fdrdecomp-ss-ssl}) is decomposed as follows:
    \begin{align}
    \underbrace{\mathbb{P}_{\Ds_{\Sh}}\{ e = 0 \} }_{\text{(E)}} =
        \underbrace{\Prob_{\Ds_{\Sh}}\{ e=0, \eh=1 \} }_{\text{FER}}
        - \underbrace{\Prob_{\Ds_{\Sh}}\{ e=1, \eh=0 \}}_{\text{FNER}}
        + \underbrace{\Prob_{\Ds_{\Sh}}\{ \eh=0 \}}_{\text{NER}}.
        \label{eq:fdrdecomp}
    \end{align}
    \vspace{-2ex}
\end{lemma}
Here, the first two terms are related to the entailment label estimation error and the last term is the approximate FDR-E using pseudo-labels. 
% As three terms are inter-related, we choose to control the FER term to control the FDR-E via conformal prediction in the following section. 
As three terms are inter-related, we choose to control the FER term to control {(E) in (\ref{eq:fdrdecomp-ss-ssl})} via conformal prediction in the following section. 

\vspace{-1ex}
\subsubsection{Pseudo-labeling via Conformalized Entailment Set Learning}
\vspace{-1ex}

\sgsemis leverages the PAC conformal prediction for the entailment label estimation to control the mislabeling error. 
Specifically, we estimate the true entailment set function $E_{\text{true}}$ via an estimated entailment set $\Eh$ using $\Z_E$, where we use the entailment scoring function $f_E$ as a scoring function, \ie $\Eh(\y) \coloneqq \{ \y' \in \Ys \mid f_E(\y', \y) \ge \tau_E \}$.
Here, the corresponding loss $\ell(\hat{E}, \x, \y, e) \coloneqq \mathbbm{1}( e=0 \wedge G(\x) \in \hat{E}(\y))$ is a monotonically non-increasing function with respect to $\tau_E$, so we can use the PAC conformal set learning algorithm.
Given a desired risk $\epsilon_E$ and confidence $\delta_E$ level, the corresponding algorithm $\As_\text{FER}$ (\ie \autoref{alg:es}) returns the estimated entailment set function $\hat{E}$ which controls the \textit{false entailment rate} (FER) of pseudo-labeled examples $\Rs_{\text{FER}}(\Eh) \coloneqq \Prob_{\Ds_{\Sh}}\{ e=0 \wedge G(\x) \in \Eh(\y) \}$ with the following PAC guarantee, where the probability is taken over training examples from $\mathcal{D}_{\hat{S}}$.
\begin{align}\label{eq:esguarantee}
\mathbbm{P} \{ \Rs_{\text{FER}}(\Eh) \le \ep_E \} \geq 1-\delta_E.
\vspace{-1ex}
\end{align}

\vspace{-1ex}
\subsubsection{FDR-E Bound}
\vspace{-1ex}
We then bound the FDR-E for semi-supervised learning, \ie (E) in (\ref{eq:fdrdecomp-ss-ssl}), via the PAC guarantee by the conformal set learning on $\Z_E$ and the binomial tail bound on $\Z_E$ and $\Z_U$. 
In particular, the {FER} is upper-bounded by $\ep_E$, the FNER is lower-bounded by the binomial tail bound using $\Z_E$, and NER is upper-bounded by the binomial tail bound using $\Z_U$. 
{These bounds hold with high probability, and are therefore combined via a union bound, as in the following lemma.}
% These bounds hold with high probability so are combined together via a union bound, as in the following lemma. 
See \autoref{apdx:proof:lem:fdrbound} for {a proof}.
\begin{lemma}\label{lem:fdrbound}
    Let
    $\hat\Z_E \coloneqq \{ (\x, \y, e) \in \Z_E \mid \Sh(\x) \neq \texttt{IDK} \}$
    and
    $\hat\Z_U \coloneqq \{ (\x, \y) \in \Z_U \mid \Sh(\x) \neq \texttt{IDK} \}$.
    For any $G$, $\mathcal{D}$, $\Vs$, and $\Sh$, if $\Eh \coloneqq \As_{\text{FER}}(\hat\Z_{E})$ satisfies $\Prob_{{\hat{\Z}_E}} \{ \Rs_{\text{FER}}(\Eh) \le \ep_E \} \ge 1 - \delta_E'/2$,
    we have
    \begin{equation}
        \mathbb{P}_{\Ds}\{ e=0 \} \leq 
        \epsilon_E 
        - L_{\text{\upshape{Binom}}}(\kh; |\hat\Z_E|, \delta_E'/2 )  
        + U_{\text{\upshape{Binom}}}(\lh; |\hat\Z_U|, \delta_S' )
        \eqcolon U_{\text{SSL}}
        \label{eq:fdr-e-semi}
    \end{equation} 
    with probability at least $1 - \delta_E' - \delta_S'$,
    where the probability is taken over $\Z$.
    Here, $\kh \coloneqq \sum_{(\x, \y, e) \in \hat{\Z}_E} \mathbbm{1}( e=1 \wedge G(\x) \notin \Eh(\y) )$
    and
    $\lh \coloneqq \sum_{(\x, \y) \in \hat{\Z}_U} \mathbbm{1}( G(\x) \notin \Eh(\y) )$.
\end{lemma}
Notably, each of three bounds holds over a conditional distribution $\Ds_{\Sh}$, but \autoref{lem:fdrbound} relaxes this to an unconditional distribution $\Ds$ for our final FDR-E guarantee.

\para{Optimizing the FDR-E Bound (\ref{eq:fdr-e-semi}).}
\autoref{lem:fdrbound} introduces a hyper-parameter $\ep_E$, which controls a trade-off between the FER and other terms. To find a best trade-off, we optimize $\ep_E$ to minimize the upper bound (\ref{eq:fdr-e-semi}) among $Q$ candidates of $\ep_E$ via $\As_{\text{$U_\text{SSL}\text{-Opt}$}}$, described in \autoref{alg:opt_u_ssl}. This optimization algorithm can find a tighter FDR-E bound, as in the following lemma. See \autoref{sec:proof:lemma:fdr-e-bound-optimized} for a proof.
% {\color{red}
% \begin{lemma}
% \label{lemma:fdr-e-bound-optimized}
%     Define $\hat{\Z}_E, \hat{\Z}_U, \hh, \lh$, and $\Eh$ as in \autoref{lem:fdrbound}. 
%     For any $G, \Ds, \Vs, \Sh$, and $Q$, we have
%     \begin{equation}
%         \mathbb{P}_{\Ds_{\Sh}}\{ e=0 \} \leq 
%         \hat{\epsilon}_E 
%         - L_{\text{\upshape{Binom}}}(\kh; |\hat\Z_E|, \delta_E'/2 )  
%         + U_{\text{\upshape{Binom}}}(\lh; |\hat\Z_U|, \delta_S' )
%         \eqqcolon U_\text{SSL}
%         \label{eq:fdr-e-semi}
%     \end{equation} 
%     with probability at least $1 - \delta_E' - \delta_S'$,
%     where the probability is taken over $\Z$ and 
%     $\hat{\ep}_E = \As_{\text{FDR-E-Opt}}(\hat{\Z}_E, \hat{\Z}_U)$. 
% \end{lemma}
% }

\begin{lemma}
\label{lemma:fdr-e-bound-optimized}
    Let $U_{\text{SSL}}$ be as in (\ref{eq:fdr-e-semi})
    and $\Qs$ be the $Q$ candidates of $\ep_E$.
    % Let $
    % U_{\text{SSL}}^{\text{OPT}} = 
    % \text{min}_{\epsilon_E} U_{\text{SSL}}$
    % be the smallest bound of (\ref{eq:fdr-e-semi}) {\color{red}, which is an optimization problem for $\ep_E$.} 
    Then, we have
    \begin{equation}
        \mathbbm{P}_{\Ds} \{ e=0 \} \leq U_{\text{SSL}}^{\text{OPT}}
        \coloneqq 
        \min_{\epsilon_E \in \Qs} U_{\text{SSL}}
        \vspace{-1ex}
    \end{equation} 
    with probability at least $1 - \delta_E'/Q - \delta_S'/Q$, where the probability is taken over $\Z$. 
\end{lemma}

Note that for semi-supervised learning, the upper bound of (B), (C), (D), and (E) in (\ref{eq:fdrdecomp-ss-ssl}) should be provided. The upper bound of (E) is provided in (\ref{eq:fdr-e-semi}), which we denote by $U_\text{SSL}$. 
The upper bound of (B), (C), and (D) are denoted by 
$w_{\text{SL}}, U_{\text{SL}}$, and $w_{\text{SSL}}$, respectively, each of which is computed by the binomial tail bound. 
See \autoref{alg:fdrbound_additional_delta} and the proof of \autoref{thm:guarantee_additional_delta} for details.

\vspace{-1ex}
\subsection{Neuro-selection Functions}
\vspace{-1ex}
The FDR-E bounds for both supervised and semi-supervised learning are crucial for controlling the final FDR-E of a selective generator given a selection function $\sh$. But, the choice of the selection function is critical for a good selection efficiency and here we discuss a better selection function than the standard one, \ie $\sh(\x) \coloneqq \mathbbm{1}\( f_M(\x, G(\x)) \ge \tau_S \)$ 
for $\tau_S \in \realnum_{\ge 0}$.
In particular, 
certified selective classification \cite{geifman2017selective} considers the single-threshold indicator function using the maximum likelihood as the confidence rate function. 
For the language generation, 
the conditional probability of the answer $\hat{\y}$, \ie $f_{M_1}( \x, \hat{\y} )$,
%the inverse of the perplexity score $f_M( \x, \hat{\y} ) \coloneqq \{ p_M(\hat{y}_1 \mid \x) \prod_{i=1}^{\mid \hat{\y} \mid}(\hat{y}_i \mid \x, \hat{\y}_{1:i-1}) \}^{\frac{1}{\mid \hat{y} \mid}}$ 
would be a natural and commonly-used candidate. 
However, as it is known to be poorly calibrated \cite{zhao2022calibrating}, an alternative would be a self-consistency score, \ie $f_{M_2}(\x, G(\x)) \coloneqq \frac{1}{K} \sum_{k=1}^K f_E(\tilde{\y}_k, G(\x))$, where $ \tilde{\y}_k$ are generated answers with the same question $\x$ but different random seeds. 
It is empirically shown that the self-consistency score properly quantifies uncertainty when a language model is uncertain of an answer \cite{manakul2023selfcheckgpt}. 
The importance of score calibration with respect to the true entailment relation is demonstrated in \autoref{lem:perfectcalent}, which provides the sufficient condition for the selective generation algorithm using the single-threshold indicator function (\autoref{alg:sgsingle_additional_delta}) to control the FDR-E at \emph{any} level. See \autoref{sec:proof:lem:perfectcalent} for a proof.
\begin{lemma}\label{lem:perfectcalent}
    If we have access to $E_{\text{true}}$ and $f_M$ is perfectly calibrated with respect to $E_\text{true}$, the FDR-E is monotonically non-increasing in $\tau_S$.
\end{lemma}
% {\color{purple}OLD:
% \begin{lemma}\label{lem:perfectcalent}
%     If an estimated entailment set function $\hat{E}$ well separates entailment labels as $E_{\text{true}}$, and $f_M$ is perfectly calibrated with respect to $\hat{E}$, the FDR-E is monotonically non-increasing in $\tau_S$.
% \end{lemma}
% }

However, as \cite{zhao2022calibrating} points out, calibrating the language scoring function remains an uneasy task, os it is still an active research area. 
Therefore, we propose a general class of selection functions, \textit{neuro-selection functions}, which is the multiple-threshold indicator function using possibly learnable feature map $\Phi: \x \mapsto \realnum^{v}$ as follows:
$
    \sh(\x; \Phi, \W, \b) \coloneqq \wedge_{i=1}^u (\W \Phi(\x))_i + \b_i \ge 0,
$
where $\W \in \realnum^{u \times v}$ and $\b \in \realnum^{u \times 1}$ are linear proejction and bias terms, respectively.
In this paper, we only consider two specific sub-classes of neuro-selection functions, where the former reduces to learning the single-threshold selection function using a scoring function (\autoref{alg:sgsingle_additional_delta}) and the latter reduces to learning the bi-threshold selection function using two scoring functions (\autoref{alg:sgdouble_additional_delta}). 
Only the bias term $\b$ is the learnable parameter for both algorithms, where the others set as hyperparameters.
Specifically, $\W = \I_1$,  $\Phi_1(\x) = [f_M(\x, G(\x))]$, and $\b = -\tau_S$ for \autoref{alg:sgsingle_additional_delta}, while $\W = \I_2$, $\Phi_2(\x) = [f_{M_1}(\x, G(\x)) ~ f_{M_2}(\x, G(\x))]^T$, and $\b = -[\tau_{S, 1}, \tau_{S, 2}]^T$ for \autoref{alg:sgdouble_additional_delta}
if two promising scoring functions exist.
%. We set the inverse perplexity as $f_{M_1}$ and the self-consistency score as $f_{M_2}$, since 
%our goal is to generate the sequence which is not only logically consistent to the true answer but also linguistically correct.
Here, developing a selection function learning algorithm where $\W$ and $\Phi(\cdot)$ are also fully learning parameters is left as future work.
In the following section, 
we introduce our algorithm that chooses the optimal combination of scoring functions via neuro-selection functions.
%\texttt{CSGen-MS} then chooses the optimal selection function among candidate scoring functions, which is the one returning the tighest FDR-E bound computed by \autoref{alg:fdrbound_additional_delta}.

%\subsection{Selective Entailing-Generation Algorithm}
\vspace{-1ex}
\subsection{Semi-Supervised Selective Generator Learning Algorithm with Neuro-Selection}
\vspace{-1ex}

\sgsemis is a semi-supervised learning algorithm for certified selective generation, which fully exploits unlabeled data in learning a selection function via certified pseudo-labeling and uses a neuro-selection function for choosing an optimal combination of scoring functions.
In particular, \sgsemis solves the following optimization problem over selective generators $\Hs$ such that $\Sh$ closely satisfies the equality in the constraint, 
as described in \autoref{alg:nsegen_additional_delta}:
%for each candidate class of selection functions:
\begin{equation}
\vspace{-1ex}
    \As_{\texttt{SGen}^\texttt{Semi}}:\quad {\operatorname{{find}}}_{\Sh \in \Hs} ~\Sh
    \quad \text{subj. to} \quad 
    w_\text{SL} U_\text{SL} + w_\text{SSL} U_\text{SSL}^{\text{OPT}}
    \le \ep_S,
    \label{eq:optforalg}
\end{equation}
Here, $\Sh \in \Hs$ has a selection function $\sh(\x; \Phi_2(\x), \texttt{diag}(\w), \b)$, where
$\w \in \{[1, 0]^T, [0, 1]^T, [1, 1]^T\}$ and $\b \in \realnum_{\le 0}^2$.
Note that 
\sgsemi returns an additional term $\hat{U}$, which is the FDR-E bound given the selective generator $\hat{S}$ (\ie \autoref{alg:fdrbound_additional_delta}) and 
informs the infeasibility of the optimization. 
The proposed \autoref{alg:nsegen_additional_delta} satisfies the following controllability guarantee.
See \autoref{sec:proof:thm:guarantee} for a proof.
\begin{theorem}\label{thm:guarantee_additional_delta}
    $\As_{\texttt{SGen}^\texttt{Semi}}$ satisfies the following controllable guarantee on the \upshape{FDR-E}, \ie
        % \begin{multline}
        %     \mathbbm{P} \Big\{ 
        %         \mathbbm{P}\{ 
        %             G(\x) \notin E_{\text{true}}(\y) \mid \hat{S}(\x) \neq \texttt{IDK} 
        %         \} \leq 
        %         \epsilon_S \mathbbm{1}(\hat{U} \leq \epsilon_S) + \hat{U} \mathbbm{1}(\hat{U} > \epsilon_S) 
        %     \Big\} \geq 1 - {\delta},
        %     \label{eq:finalbound}
        % \end{multline}
        \vspace{-1ex}
        \begin{equation}
            \mathbbm{P} \Big\{ 
                \mathbbm{P}\{ 
                    G(\x) \notin E_{\text{true}}(\y) \mid \hat{S}(\x) \neq \texttt{IDK} 
                \} \leq 
                \hat{U}
            \Big\} \geq 1 - {\delta},
            \label{eq:finalbound}
            \vspace{-1ex}
        \end{equation}
    where the inner and outer probabilities are taken over $(\x, \y, e, v) \sim \mathcal{D}$ and $\Z \sim \mathcal{D}^n$, respectively, and $(\hat{S}, \hat{U}) \coloneqq \mathcal{A}_{\texttt{SGen}^\texttt{Semi}}(\Z)$. 
    Here, $\delta \coloneqq \delta_W + \delta_S + \delta_E$ is a desired confidence level, where
    $\delta_W$ is for
    the upper bounds on $w_\text{SL}$ and $w_\text{SSL}$, $\delta_S$ is for (C) in (\ref{eq:fdrdecomp-ss-ssl}) and the NER, and $\delta_E$ is for the FER and FNER.
\end{theorem}
Here, $\mathcal{A}_{\texttt{SGen}^\texttt{Semi}}$ is \emph{controllable} in the sense that it upper-bounds the FDR-E of a learned selective generator to a desired level $\epsilon_S$ or at least to a minimum achievable level $\Uh$ with confidence $\delta$. 

%Recall that under some conditions, perfect calibration is a sufficient condition for the selective generation algorithm using the single-threshold indicator function as selection function (\autoref{alg:sgsingle_additional_delta}) to control FDR-E at any level (\autoref{lem:perfectcalent}).
%However, since calibrating language scoring functions is difficult, we instead consider a general class of selection functions called \textit{neuro-selection functions}.
\vspace{-1ex}
\section{Experiments}\label{sec:eval}
\vspace{-1ex}
We demonstrate the efficacy of our methods in controlling the FDR-E on pre-trained GLMs under various setups. We use two GLMs, GPT-3.5-Turbo and Alpaca-7B, alongside the Natural Questions (NQ) dataset to annotate entailment labels for question-answer pairs. Details on model configurations, datasets, and additional experimental results can be found in \autoref{sec:apdx:expsetup} and \autoref{sec:apdx:addexps}.

%We demonstrate the efficacy of the proposed method \texttt{NSEGen} compared to three baselines on open-source and closed source GLMs. 

%\subsection{Setup}
% \para{Models and Datasets.} 
% {\color{red} [MJ: No need?]}
% We use two GLMs, \ie \textit{GPT-3.5-Turbo} and \textit{Alpaca-7B}, for language generation, and use the Natural Questions (NQ) dataset \cite{kwiatkowski2019natural} for each GLM to annotate entailment labels for each question, answer, and generated answer pair. See \autoref{sec:apdx:modelsdatasets} for details.

%% main results

\begin{table*}
\centering

\caption{Comparison results of semi-supervised methods. Here, 
$|\Z_U|=10K$ for GPT-3.5-turbo and Alpaca-7B.
The best results are highlighted in \textbf{bold}
and 
results from methods that do not satisfy desired FDR-E guarantees in learning are \underline{underlined}.
}
\label{tab:alpaca7B:semi-supervised}

\def\arraystretch{1}
\setlength{\tabcolsep}{2pt}
\footnotesize
\resizebox{\textwidth}{!}{ % Resizes the table to fit within the text width
\footnotesize
\begin{tabular}{cc|ccccc|ccccc}
\toprule
\multicolumn{2}{c|}{Models} & \multicolumn{5}{c|}{GPT-3.5-turbo} & \multicolumn{5}{c}{Alpaca-7B} 
\\
\midrule
% GPT-3.5-turbo
\multicolumn{2}{c|}{\multirow{2}{*}{Methods}} & \multicolumn{2}{c}{Heuristic} & \multicolumn{3}{c|}{Certified} & \multicolumn{2}{c}{Heuristic} & \multicolumn{3}{c}{Certified}
\\
\cmidrule(lr){3-4} \cmidrule(lr){5-7} \cmidrule(lr){8-9} \cmidrule(lr){10-12}
\multicolumn{2}{c|}{} & \sgpl & \sgpfl & \sgem & \sgseminoms & \sgsemi
% Alpaca-7B
& \sgpl & \sgpfl & \sgem & \sgseminoms & \sgsemi
\\
\midrule
\midrule
\multirow{2}{*}{$f_{M_1}$} & FDR-E
% GPT-3.5-turbo
& $0.0958$ & $0.0283$ & $\underline{0.1338}$ & $\underline{0.0609}$ & $0.1589$
% Alpaca-7B
& $0.0231$ & ${0.0068}$ & $\underline{0.0359}$ & $\underline{0.0359}$ & $0.0685$
\\
& efficiency 
% GPT-3.5-turbo
& $0.4189$ & $0.1719$ & $\underline{0.5495}$ & $\underline{0.2829}$ & $0.7334$
% Alpaca-7B
& $0.0915$ & ${0.0332}$ & $\underline{0.1580}$ & $\underline{0.1580}$ & $0.3173$
\\
\midrule
\multirow{2}{*}{$f_{M_2}$} & FDR-E
% GPT-3.5-turbo
& $0.1839$ & $0.2002$ & $\underline{0.0914}$ & $0.1785$ & $0.1589$
% Alpaca-7B
& $0.0698$ & $0.0732$ & $\underline{0.0549}$ & $0.0698$ & $0.0685$
\\
& efficiency
% GPT-3.5-turbo
& $0.7911$ & $0.8183$ & $\underline{0.5332}$ & $0.7769$ & $0.7334$
% Alpaca-7B
& $0.3207$ & $0.3390$ & $\underline{0.2563}$ & $0.3200$ & $0.3173$
\\
\midrule\midrule
% \multicolumn{2}{c|}{average FDR-E}
% % GPT-3.5-turbo
% & $0.055$ & $0.065$ & $0.085$
% % Alpaca-7B
% & $0.02475$ & $0.0268$ & $0.0244$
% \\
\multicolumn{2}{c|}{average efficiency}
% GPT-3.5-turbo
%& $0.4950$ & $0.7375$ & $0.4825$ & $0.8200$
& $0.6050$ & $0.4951$ & $-$ & $-$ & $\mathbf{0.7334}$
% Alpaca-7B
%& $0.1481$ & $0.1600$ & $0.1498$ & $0.1532$
& $0.2061$ & $0.1861$ & $-$ & $-$ & $\mathbf{0.3173}$
\\
\bottomrule
\end{tabular}
} % Ends resizebox
\end{table*}

\para{Methods.} We consider two heuristic semi-supervised algorithms, \sgpls and \sgpfls (\autoref{alg:selgenpsl}) and an unsupervised learning algorithm \cite{geifman2017selective} \sgems (\autoref{alg:selgenem}) as baselines to show the efficacy of our certified semi-supervised method \sgsemis (\autoref{alg:nsegen_additional_delta}).
\sgpls and \sgpfls exploit the unlabeled data by pseudo-labeling textual entailment based on a threshold as a hyperparameter without any guarantee on mislabeling error.
\sgpfls additionally filters out a pseudo-labeled sample if its entailment score is below a specific threshold.
\sgems is a certified unsupervised method that takes the EM metric for measuring the correctness.
We also report results on \sgseminoms (\autoref{alg:sgsingle_additional_delta}) for two different scoring functions ,$f_{M_1}$ and $f_{M_2}$, used in \sgsemi. 
\sgseminomss is a certified semi-supervised learning algorithm using a single-threshold indicator function given a scoring function. 
We also take \sgsups (\autoref{alg:selgenel}) as a baseline, since it is a direct modification of \cite{geifman2017selective} to the language generation problem.

% \begin{itemize}
%     \item supervised
%     \begin{itemize}
%     \item (certified) \texttt{SGen-EM}$\to$ \sgem
%     \item (certified) \texttt{SGen-EL}$\to$ \sgsup
%     \item (certified) \texttt{CSGen-Sup}$\to$ \sgsemisup \SP{not good, but fine as we will have this only in the appendix.}
%     \end{itemize}
%     \item semi-supervised
%     \begin{itemize}
%     \item (heuristic) \texttt{SGen-PL}$\to$ \sgpl
%     \item (heuristic) \texttt{SGen-PFL}$\to$ \sgpfl
%     \item (certified) \texttt{CSGen}$\to$ \sgseminoms
%     \item (certified) \texttt{CSGen-MS}$\to$ \sgsemi
%     \end{itemize}
% \end{itemize}

% f_M
\para{Scoring Functions.} We use the conditional probability of an answer as $f_{M_1}$ and the self-consistency score \cite{manakul2023selfcheckgpt} as $f_{M_2}$, since our goal is to generate the sequence which is not only logically consistent to the true answer but also linguistically correct.

%\begin{itemize}[leftmargin=*,topsep=0pt,itemsep=0.3ex,partopsep=1ex,parsep=0ex]
% \item \textbf{SGen-EM} \cite{geifman2017selective}: Selective generation with exact match (\autoref{alg:selgenem})
% \item \textbf{SGen-EL}: Selective generation with entailment labels (\autoref{alg:selgenel}) 
% \item \textbf{SGen-ES}: Selective generation with an entailment set (\autoref{alg:sgsingle})
% \item \textbf{NSEGen} (ours): neuro-selective entailing-generation (\autoref{alg:nsegen})
%\end{itemize}
% parameters

\para{Control Parameters.}
To control an FDR-E, we use two user-specified parameters $(\ep, \delta)$, where we use $(0.25, 0.02)$ unless specified. 
For our methods (\ie \sgsemi, \sgseminoms, and \sgsemisup), we have five control parameters ($\ep_S, \delta_S, \delta_E, \delta_W$), where we maps as follows: $\ep_S = \ep$, $\delta_S = (\delta-\delta_W)/2, \delta_E = (\delta-\delta_W)/2,  \delta_W = 10^{-5}$.
For other methods without using entailment sets, \autoref{alg:selgenel}, \autoref{alg:selgenpsl}, and \autoref{alg:selgenem}, we use $\ep$ and $\delta$ accordingly.
Additionally, we use $Q=5$ for \autoref{alg:opt_u_ssl}.

% \texttt{CSGen} and \texttt{CSGen-NS} have 4 user-specified parameters ($\ep_S, \ep_E, \delta_S, \delta_E$). 
% For fair comparison with other methods (\ie \texttt{SGen-PL} and \texttt{SGen-PFL}), we set $(\ep, \delta) = (\ep_S, \delta_S+\delta_E) = (0.25, 0.02)$.
% We use $\ep_E$ for finding entailment set and $\ep_S-\ep_E$ for finding a final selective generator, guaranteeing FDR-E is bounded by $\ep_S=\ep$ as sames as baselines.
% \SP{update}

%In our methods, which have entailment set (\textbf{SGen-ES} and \textbf{NSEGen}), we have 4 hyper-parameters($\ep_S, \ep_E, \delta_S, \delta_E$). Thus, for the purpose of equal comparison with baselines (\textbf{SGen-EM} and \textbf{SGen-EL}), we set $(\ep, \delta)$ as $(\ep_S, \delta_S+\delta_E)$ individually and  we tune $\ep_E$ to find reasonable performance.

%Also, our method can make use of another $f_M$ score function at the same time. Since the perplexity for the generator implies a kind of confidence, it does not mean the perplexity can well separate the domain. We address new scoring function $f_{M2}$ inspired by SelfCheckGPT\cite{manakul2023selfcheckgpt}, which measures the consistency(the mean of entailment score).
\begin{figure*}
  \subfigure[\sgseminomss on Alpaca7B]{
    \includegraphics[width=0.43\linewidth]{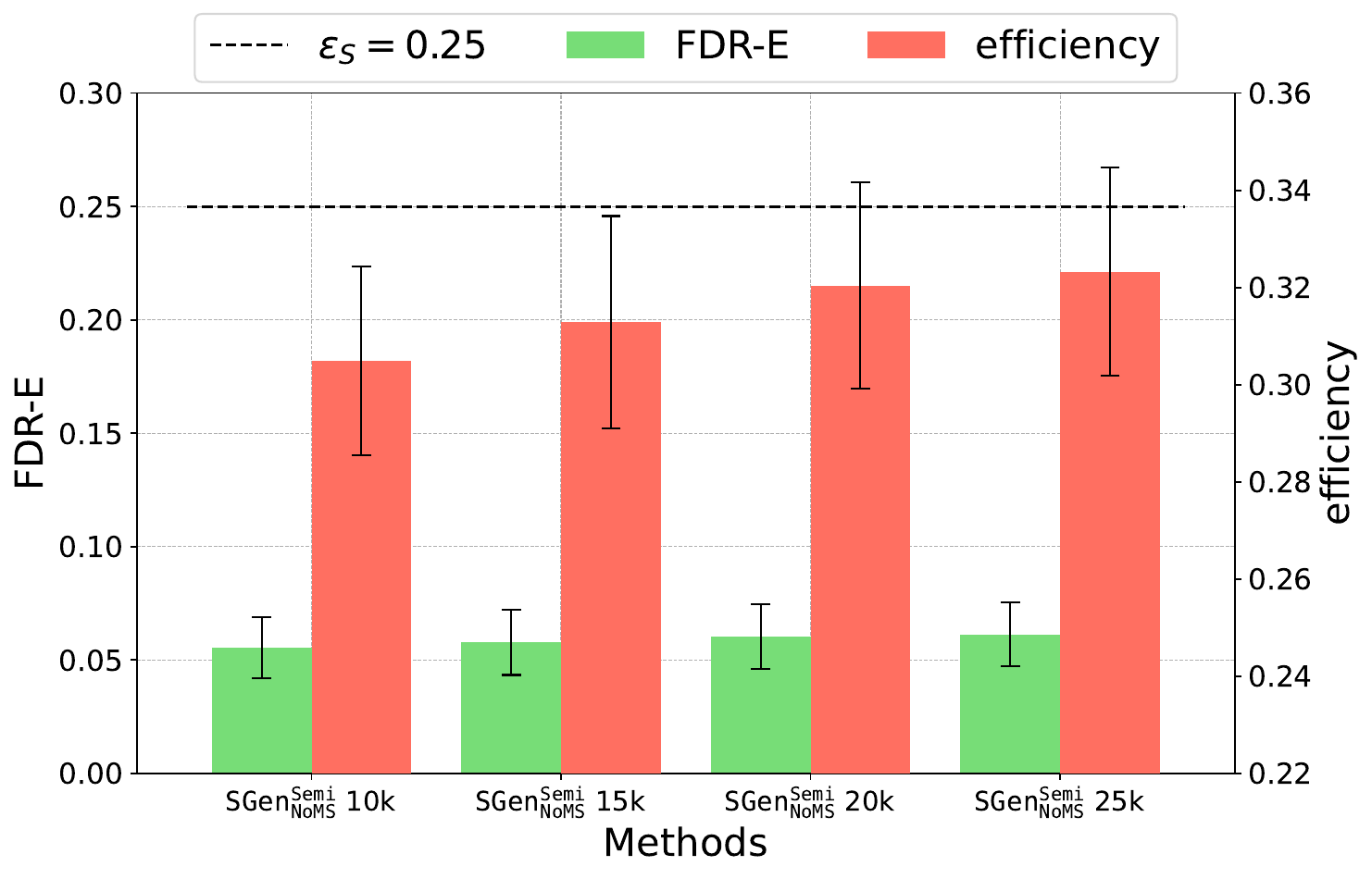}
    \label{fig:alpaca7B:CSGen:quantity:plot}
  }
  \hfill
  \subfigure[\sgsemis on Alpaca7B]{
    \includegraphics[width=0.43\linewidth]{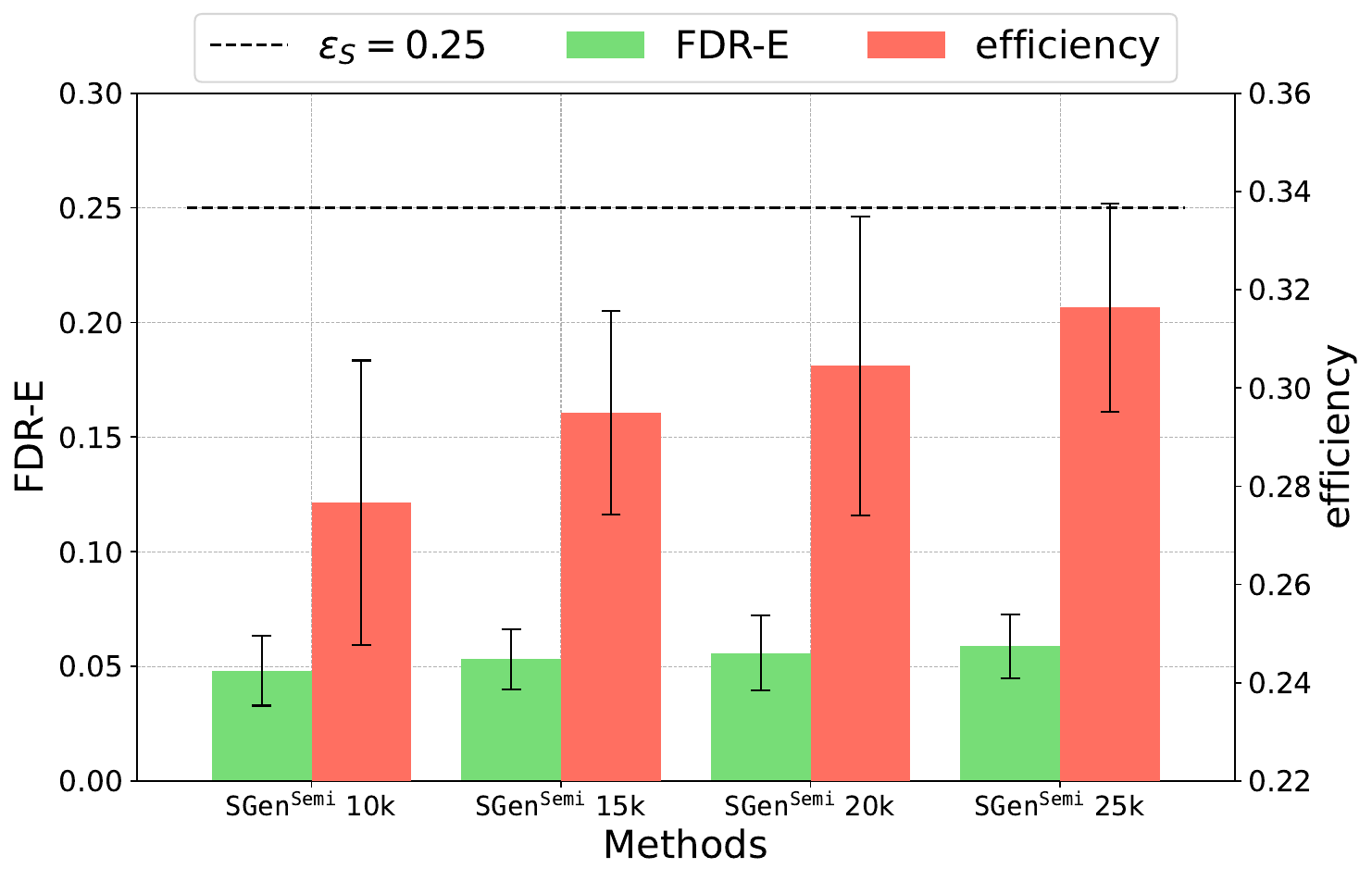}
    \label{fig:alpaca7B:CSGen-NS:quantity:plot}
  }
  \subfigure[\sgseminomss on GPT-3.5-turbo]{
    \includegraphics[width=0.43\linewidth]{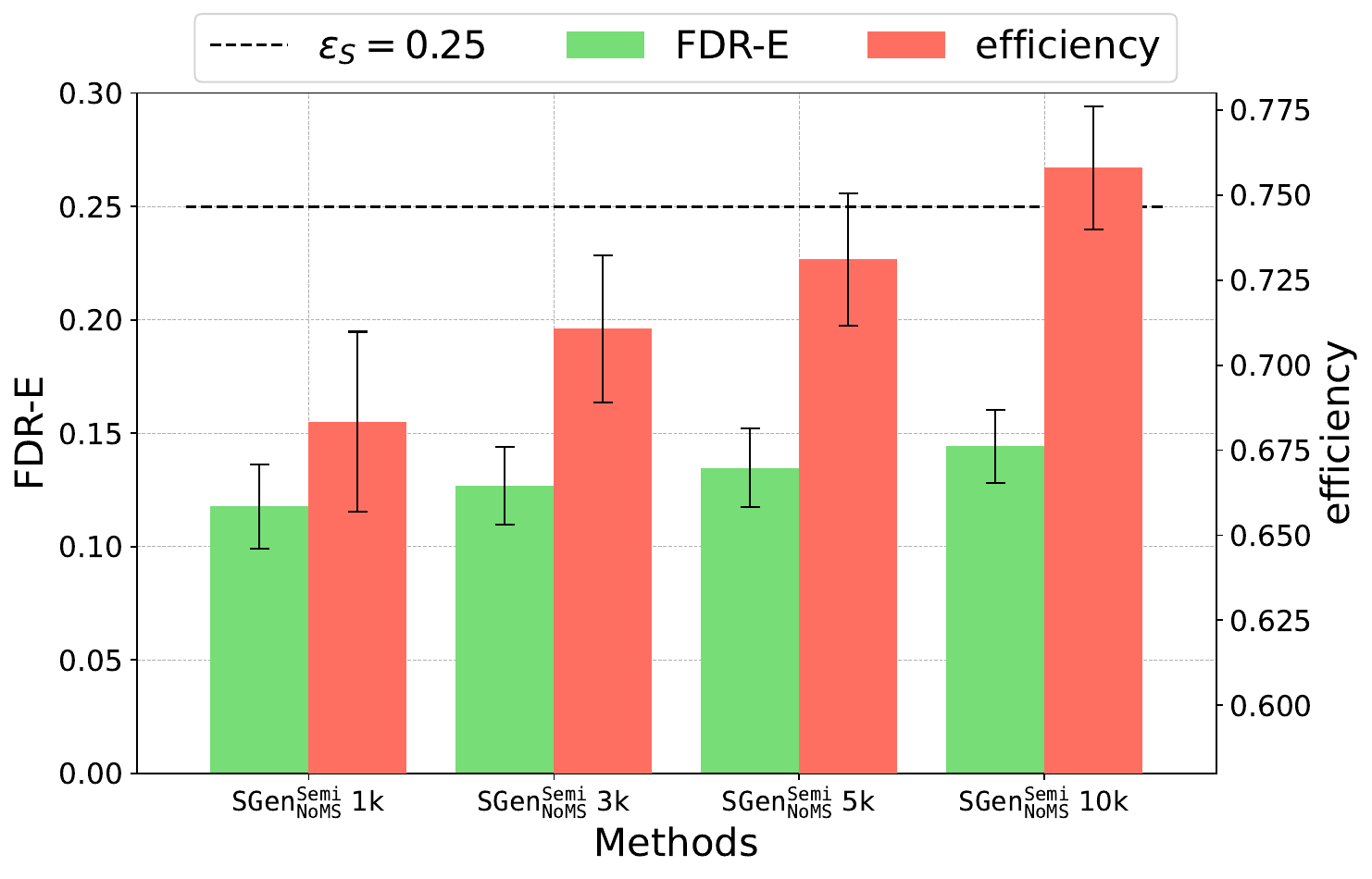}
    \label{fig:gpt-3.5:CSGen:quantity:plot}
  }
  \hfill
  \subfigure[\sgsemis on GPT-3.5-turbo]{
    \includegraphics[width=0.43\linewidth]{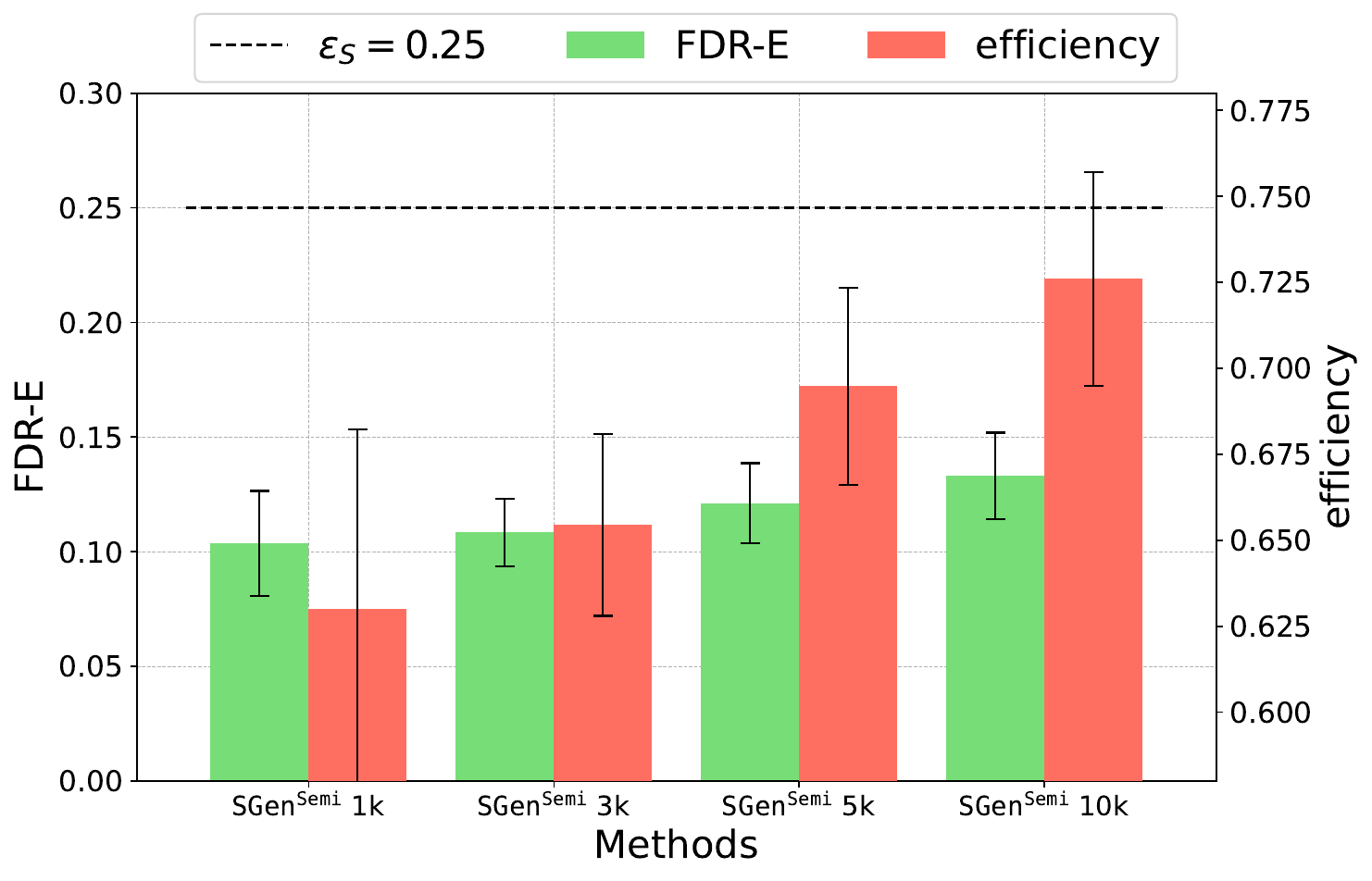}
    \label{fig:gpt-3.5:CSGen-NS:quantity:plot}
  }
  \vspace{-1ex}
  \caption{Efficiency results over different numbers of unlabeled samples. 
  (a) and (b) use \sgseminomss with $f_{M_2}$ score. (c) and (d) use \sgsemis that has neuro-selection function. Both methods show increasing performance as more unlabeled samples $\Z_U$ are used.
  For each experiment, the values were measured after averaging 10 random splits and an error bar means standard deviation.
  }
  
  \label{fig:quant}
  \vspace{-3ex}
\end{figure*}

\begin{table*}[tb!]
\vspace{-2.5ex}
  \centering
  \footnotesize
  \setlength{\fboxsep}{0pt}
  \def\arraystretch{0.5}
  \setlength{\tabcolsep}{2pt}
  
  \caption{Qualitative results by Alpaca7B. } 
  \label{tab:alpaca7B:qualitative}

  \begin{tabular}{c|p{4.8cm}||p{4.8cm}}
    \toprule

    \makecell{\textbf{Question $\x$}}
    &
    {Who is the actor who plays Draco Malfoy?}&
    {When did the movie Benjamin Button come out?}
    \\
    \midrule

    \textbf{Correct Answer $\y$}
    &
    \makecell{Thomas Andrew Felton plays Draco\\Malfoy in the Harry Potter movies.}&
    \makecell{The movie Benjamin Button\\come out December 25, 2008}
    \\
    \midrule

    % standard baseline
    \makecell{Generated Answer $G(\x)$}
    &
    \makecell{The actor who plays Draco Malfoy is\\Tom Felton. ({\color{ForestGreen}{correct}})}&
    \makecell{The movie The Curious Journey\\of Benjamin Button was\\released in 2008. ({\color{ForestGreen}{correct}})}\\
    \midrule
    
    \makecell{\sgems \cite{geifman2017selective}}
    &
    \makecell{{\color{red}rejected}}
    &
    \makecell{{\color{red}rejected}}
    \\
    \midrule
    
    \makecell{\sgsemi (ours)}
    &
    \makecell{{\color{ForestGreen}{accepted}}}
    &
    \makecell{{\color{ForestGreen}{accepted}}}\\

    \bottomrule
  \end{tabular}
  \vspace{-2ex}
\end{table*}

\para{FDR-E Guarantee and Efficiency.}
%\para{does our method achieve a desired guarantee, while achieving the selection efficiency?}
As can be seen in \autoref{tab:alpaca7B:semi-supervised}, our method \sgsemis can overall achieve desired FDR-E guarantees with better efficiency compared to baselines.
Depending on the quality of scoring functions (\eg $f_{M_1}$), our variation \sgseminomss may not find a selective generator that satisfies a desired FDR-E (denoted in the underlined FDR-E). 
The heuristic methods, \ie \sgpls and \sgpfl, do not provide theoretical guarantees on FDR-E. 
%As expected, the naive selective generation (\texttt{SGen-EM}, \autoref{tab:alpaca7B:fully-supervised}), which uses exact match, poorly performs as it cannot capture the semantic correctness, while we do via entailment sets. 
%Considering any two unknown $f_M$ scoring functions, we can utilize them by multiple-$\tau$ strategy. (\autoref{alg:nsegen})
%\para{what's the qualitative results of our method?}
In \autoref{fig:method_overview} and \autoref{tab:alpaca7B:qualitative}, we can correctly predict even with the complicated answers, e.g., which have many equivalent words, because we do not rely on the EM metric. 
%Also, in the second example in \autoref{tab:gpt-3.5:qualitative}, an answer that are lexically distinct can entail the true answer by our method. 
%We can consider other scores, which could make the domain more separable, and use entailment labels.
% \SP{summarize results on \autoref{fig:boxplot:gpt3.5} -- we did random experimentsXXX }
We conducted 100 random experiments for each method to show how well FDR-E is bounded under a desired FDR-E.
As shown by the green boxes In \autoref{fig:boxplot:gpt3.5}, which are successfully bounded under $\ep_S=0.25$, we can see that the FDR-E for a learned selective generator is well controlled below $\ep_S$ under the test environment.
Among the certified methods with theoretical guarantees, results appear to align well with the expected theoretical basis.

%Interesting thing is that there is no result such that there is a data rejected by NCG but accepted by SGen-EL or SGen-EM in GPT-3.5.

%There are no accepted results in SGen-EM, which suggests that EM metric is not adequate for selective generation task.

\para{Why Entailment Labels.}
%\para{why do we need entailment labels?}
As expected and can be seen in \autoref{tab:alpaca7B:fully-supervised} by comparing \sgems and \sgsup, a metric like EM cannot measure correctness correctly. Unlike classification, generative tasks can have infinite number of true answers so it is not likely to have exact match. Instead, entailment labels provide semantic correctness, so \sgsups can perform better and more efficient than \sgem.
%to approximate model's generated sequence is relevant to the true answer.

%\para{Why Entailment Sets.}
\para{Why Semi-Supervised Learning.}
%\para{why do we need an entailment set?}
We observe that our semi-supervised learning for selective generation is effective. 
In particular, 
the fully supervised methods in \autoref{tab:alpaca7B:fully-supervised} achieves the efficiency of $0.7535$ and $0.2959$ for GPT-3.5 and Alpaca-7B, respectively, with the entire labeled samples $\Z_E$ (when they satisfy a $\ep$-FDR-E guarantee). Compared to these, the proposed semi-supervised method \sgsemis \autoref{tab:alpaca7B:semi-supervised} achieves the efficiency of $0.7334$ and $0.3173$ for GPT-3.5 and Alpaca-7B, respectively, by only using $75\%$ of labeled examples. 
% more samples
Additionally, we observe that more unlabeled samples are beneficial to achieving better efficiency as can be seen in \autoref{fig:quant}.
This implies that 
if we can approximate the entailment set well and the size of $\Z_U$ is enough, we can enjoy our  certified pseudo-entailment labeling by the semi-supervised learning even with small $\Z_E$.

% In \autoref{tab:alpaca7B:semi-supervised} and \autoref{tab:alpaca7B:fully-supervised}, For GPT-3.5, the best efficiency of semi-supervised methods is $0.8200$ compared to $0.9075$ fully-supervised methods.
% We can see heuristic semi-supervised methods can achieve nearly fully-supervised methods even with smaller calibration set ($0.75 |\Z_E^\text{cal}|$), and the certified method (\texttt{CSGen-NS}) performs better than heuristic methods in average. 
% We can also see the performance becomes much better as $|\Z_U|$ gets bigger in \autoref{fig:quant}.
% Theses results mean that if we can approximate the entailment set well and there is enough $\Z_U$, we can enjoy the pseudo-entailment labels by the semi-supervised learning even with small $\Z_E$.
% The greatest weakness of solving selective generation tasks with entailment is the expensive cost of labeling. It can be mitigated if we use the better entailment network and much more calibrated $f_M$ scores.

\para{Why Neuro-Selection.}
%\para{why do we need feature selection?}
It is hard to manually find a well calibrated scoring function. But, given multiple scoring functions, a neuro-selection function learns to choose right scoring functions that achieves a desired FDR-E and maximizes selection efficiency. 
This is empiricially validated in \autoref{tab:alpaca7B:semi-supervised}, as \sgsemis is better on average efficiency.
\vspace{-2ex}
\section{Conclusion} \label{sec:conc}
\vspace{-2ex}
We propose selective generation, a generalized version of \cite{geifman2017selective} for GLMs 
to handle semantic correctness between two structured answers. 
To this end, 
we leverage logical entailment to define a new entailment-based FDR (FDR-E) metric. As obtaining entailment labels are expensive, we propose novel semi-supervised learning for selective generation by using entailment sets as a pseudo-labeling function. To enhance the low selective efficiency due to inefficient scoring functions,  
we propose neuro-selection functions for effectively optimizing scoring functions for better selective efficiency and the FDR-E guarantee.
The efficacy of our proposed algorithms \sgsemis and \sgsups are theoretically and empirically justified. 

\para{Limitations.} 
Our algorithm needs the i.i.d. assumption for a correctness guarantee, which can be violated in practical situations. 
We leverage expensive entailment labels, where the labels are obtained by considering logical entailment between a true answer and a generated answer. This limitation is partially mitigated by proposing the semi-supervised method to propagate entailment-labeled samples to samples without entailment labels.
Also, our results show the empirical FDR-E is not much closely bounded under $\ep$, especially on Alpaca7B, which implies that we may need a tighter FDR-E bound.

\clearpage
\section*{Acknowledgements}

This work was supported by Institute of Information \& communications Technology Planning \& Evaluation (IITP) grant funded by the Korea government (MSIT) (No.RS-2019-II191906, Artificial Intelligence Graduate School Program (POSTECH) (50\%); RS-2024-00457882, National AI Research Lab Project (25\%); RS-2024-00509258, Global AI Frontier Lab (25\%)).
Also, we appreciate valuable comments by NeurIPS reviewers.

\bibliographystyle{unsrt}
\bibliography{llm,sml}

\begin{thebibliography}{10}

\bibitem{radford2019language}
Alec Radford, Jeffrey Wu, Rewon Child, David Luan, Dario Amodei, Ilya Sutskever, et~al.
\newblock Language models are unsupervised multitask learners.
\newblock {\em OpenAI blog}, 1(8):9, 2019.

\bibitem{brown2020language}
Tom~B. Brown, Benjamin Mann, Nick Ryder, Melanie Subbiah, Jared Kaplan, Prafulla Dhariwal, Arvind Neelakantan, Pranav Shyam, Girish Sastry, Amanda Askell, Sandhini Agarwal, Ariel Herbert-Voss, Gretchen Krueger, Tom Henighan, Rewon Child, Aditya Ramesh, Daniel~M. Ziegler, Jeffrey Wu, Clemens Winter, Christopher Hesse, Mark Chen, Eric Sigler, Mateusz Litwin, Scott Gray, Benjamin Chess, Jack Clark, Christopher Berner, Sam McCandlish, Alec Radford, Ilya Sutskever, and Dario Amodei.
\newblock Language models are few-shot learners, 2020.

\bibitem{touvron2023llama}
Hugo Touvron, Thibaut Lavril, Gautier Izacard, Xavier Martinet, Marie-Anne Lachaux, Timothée Lacroix, Baptiste Rozière, Naman Goyal, Eric Hambro, Faisal Azhar, Aurelien Rodriguez, Armand Joulin, Edouard Grave, and Guillaume Lample.
\newblock Llama: Open and efficient foundation language models, 2023.

\bibitem{alpaca}
Rohan Taori, Ishaan Gulrajani, Tianyi Zhang, Yann Dubois, Xuechen Li, Carlos Guestrin, Percy Liang, and Tatsunori~B. Hashimoto.
\newblock Stanford alpaca: An instruction-following llama model.
\newblock \url{https://github.com/tatsu-lab/stanford_alpaca}, 2023.

\bibitem{chatgpt}
{OpenAI Team}.
\newblock {ChatGPT}.
\newblock https://chat.openai.com/, 2021.

\bibitem{vaswani2017attention}
Ashish Vaswani, Noam Shazeer, Niki Parmar, Jakob Uszkoreit, Llion Jones, Aidan~N Gomez, Lukasz Kaiser, and Illia Polosukhin.
\newblock Attention is all you need.
\newblock {\em Advances in neural information processing systems}, 30, 2017.

\bibitem{li2020dont}
Margaret Li, Stephen Roller, Ilia Kulikov, Sean Welleck, Y-Lan Boureau, Kyunghyun Cho, and Jason Weston.
\newblock Don{'}t say that! making inconsistent dialogue unlikely with unlikelihood training.
\newblock In {\em Proceedings of the 58th Annual Meeting of the Association for Computational Linguistics}, pages 4715--4728, Online, July 2020. Association for Computational Linguistics.

\bibitem{ouyang2022training}
Long Ouyang, Jeff Wu, Xu~Jiang, Diogo Almeida, Carroll~L. Wainwright, Pamela Mishkin, Chong Zhang, Sandhini Agarwal, Katarina Slama, Alex Ray, John Schulman, Jacob Hilton, Fraser Kelton, Luke Miller, Maddie Simens, Amanda Askell, Peter Welinder, Paul Christiano, Jan Leike, and Ryan Lowe.
\newblock Training language models to follow instructions with human feedback, 2022.

\bibitem{geifman2017selective}
Yonatan Geifman and Ran El-Yaniv.
\newblock Selective classification for deep neural networks.
\newblock {\em Advances in neural information processing systems}, 30, 2017.

\bibitem{vovk2005algorithmic}
Vladimir Vovk, Alex Gammerman, and Glenn Shafer.
\newblock {\em Algorithmic learning in a random world}.
\newblock Springer Science \& Business Media, 2005.

\bibitem{Park2020PAC}
Sangdon Park, Osbert Bastani, Nikolai Matni, and Insup Lee.
\newblock Pac confidence sets for deep neural networks via calibrated prediction.
\newblock In {\em International Conference on Learning Representations}, 2020.

\bibitem{bates2021distribution}
Stephen Bates, Anastasios Angelopoulos, Lihua Lei, Jitendra Malik, and Michael~I Jordan.
\newblock Distribution-free, risk-controlling prediction sets.
\newblock {\em arXiv preprint arXiv:2101.02703}, 2021.

\bibitem{gibbs2021adaptive}
Isaac Gibbs and Emmanuel Cand{\`e}s.
\newblock Adaptive conformal inference under distribution shift, 2021.

\bibitem{park2023acon2}
Sangdon Park, Osbert Bastani, and Taesoo Kim.
\newblock Acon$^2$: Adaptive conformal consensus for provable blockchain oracles, 2023.

\bibitem{bowman2015large}
Samuel Bowman, Gabor Angeli, Christopher Potts, and Christopher~D Manning.
\newblock A large annotated corpus for learning natural language inference.
\newblock In {\em Proceedings of the 2015 Conference on Empirical Methods in Natural Language Processing}, pages 632--642, 2015.

\bibitem{williams2018broad}
Adina Williams, Nikita Nangia, and Samuel~R Bowman.
\newblock A broad-coverage challenge corpus for sentence understanding through inference.
\newblock In {\em Proceedings of NAACL-HLT}, pages 1112--1122, 2018.

\bibitem{kwiatkowski2019natural}
Tom Kwiatkowski, Jennimaria Palomaki, Olivia Redfield, Michael Collins, Ankur Parikh, Chris Alberti, Danielle Epstein, Illia Polosukhin, Jacob Devlin, Kenton Lee, et~al.
\newblock Natural questions: a benchmark for question answering research.
\newblock {\em Transactions of the Association for Computational Linguistics}, 7:453--466, 2019.

\bibitem{jiang2021howcan}
Zhengbao Jiang, Jun Araki, Haibo Ding, and Graham Neubig.
\newblock {How Can We Know When Language Models Know? On the Calibration of Language Models for Question Answering}.
\newblock {\em Transactions of the Association for Computational Linguistics}, 9:962--977, 09 2021.

\bibitem{manakul2023selfcheckgpt}
Potsawee Manakul, Adian Liusie, and Mark Gales.
\newblock Selfcheckgpt: Zero-resource black-box hallucination detection for generative large language models.
\newblock In {\em The 2023 Conference on Empirical Methods in Natural Language Processing}, 2023.

\bibitem{vovk2013conditional}
Vladimir Vovk.
\newblock Conditional validity of inductive conformal predictors.
\newblock {\em Machine learning}, 92(2-3):349--376, 2013.

\bibitem{fisch2021fewshot}
Adam Fisch, Tal Schuster, Tommi Jaakkola, and Regina Barzilay.
\newblock Few-shot conformal prediction with auxiliary tasks, 2021.

\bibitem{park2022pacmeta}
Sangdon Park, Edgar Dobriban, Insup Lee, and Osbert Bastani.
\newblock {PAC} prediction sets for meta-learning.
\newblock In Alice~H. Oh, Alekh Agarwal, Danielle Belgrave, and Kyunghyun Cho, editors, {\em Advances in Neural Information Processing Systems}, 2022.

\bibitem{quach_conformal_2024}
Victor Quach, Adam Fisch, Tal Schuster, Adam Yala, Jae~Ho Sohn, Tommi~S. Jaakkola, and Regina Barzilay.
\newblock Conformal {Language} {Modeling}, June 2024.
\newblock arXiv:2306.10193 [cs].

\bibitem{mohri2023learning}
Christopher Mohri, Daniel Andor, Eunsol Choi, and Michael Collins.
\newblock Learning to reject with a fixed predictor: Application to decontextualization.
\newblock {\em arXiv preprint arXiv:2301.09044}, 2023.

\bibitem{jin_selection_2023}
Ying Jin and Emmanuel~J. Candès.
\newblock Selection by {Prediction} with {Conformal} p-values, May 2023.
\newblock arXiv:2210.01408 [stat].

\bibitem{jin_confidence_2024}
Ying Jin and Zhimei Ren.
\newblock Confidence on the {Focal}: {Conformal} {Prediction} with {Selection}-{Conditional} {Coverage}, March 2024.
\newblock arXiv:2403.03868 [math, stat].

\bibitem{mohri2024language}
Christopher Mohri and Tatsunori Hashimoto.
\newblock Language models with conformal factuality guarantees.
\newblock {\em arXiv preprint arXiv:2402.10978}, 2024.

\bibitem{cherian_large_2024}
John~J. Cherian, Isaac Gibbs, and Emmanuel~J. Candès.
\newblock Large language model validity via enhanced conformal prediction methods, June 2024.
\newblock arXiv:2406.09714 [cs, stat].

\bibitem{gui_conformal_2024}
Yu~Gui, Ying Jin, and Zhimei Ren.
\newblock Conformal {Alignment}: {Knowing} {When} to {Trust} {Foundation} {Models} with {Guarantees}, May 2024.
\newblock arXiv:2405.10301 [cs, stat].

\bibitem{gage1994new}
Philip Gage.
\newblock A new algorithm for data compression.
\newblock {\em C Users Journal}, 12(2):23--38, 1994.

\bibitem{wilks1941determination}
Samuel~S Wilks.
\newblock Determination of sample sizes for setting tolerance limits.
\newblock {\em The Annals of Mathematical Statistics}, 12(1):91--96, 1941.

\bibitem{valiant1984theory}
Leslie~G Valiant.
\newblock A theory of the learnable.
\newblock {\em Communications of the ACM}, 27(11):1134--1142, 1984.

\bibitem{papadopoulos2002inductive}
Harris Papadopoulos, Kostas Proedrou, Volodya Vovk, and Alex Gammerman.
\newblock Inductive confidence machines for regression.
\newblock In {\em European Conference on Machine Learning}, pages 345--356. Springer, 2002.

\bibitem{Park2022PAC}
Sangdon Park, Edgar Dobriban, Insup Lee, and Osbert Bastani.
\newblock {PAC} prediction sets under covariate shift.
\newblock In {\em International Conference on Learning Representations}, 2022.

\bibitem{degroot1983comparison}
Morris~H DeGroot and Stephen~E Fienberg.
\newblock The comparison and evaluation of forecasters.
\newblock {\em Journal of the Royal Statistical Society: Series D (The Statistician)}, 32(1-2):12--22, 1983.

\bibitem{zadrozny2002transforming}
Bianca Zadrozny and Charles Elkan.
\newblock Transforming classifier scores into accurate multiclass probability estimates.
\newblock In {\em Proceedings of the eighth ACM SIGKDD international conference on Knowledge discovery and data mining}, pages 694--699. ACM, 2002.

\bibitem{zhao2022calibrating}
Yao Zhao, Mikhail Khalman, Rishabh Joshi, Shashi Narayan, Mohammad Saleh, and Peter~J Liu.
\newblock Calibrating sequence likelihood improves conditional language generation.
\newblock In {\em The Eleventh International Conference on Learning Representations}, 2022.

\bibitem{demszky2018transforming}
Dorottya Demszky, Kelvin Guu, and Percy Liang.
\newblock Transforming question answering datasets into natural language inference datasets.
\newblock {\em arXiv preprint arXiv:1809.02922}, 2018.

\bibitem{chen_can_2021}
Jifan Chen, Eunsol Choi, and Greg Durrett.
\newblock Can {NLI} {Models} {Verify} {QA} {Systems}' {Predictions}?, September 2021.
\newblock arXiv:2104.08731 [cs].

\end{thebibliography}

\clearpage
\appendix
\onecolumn

\section{Discussion}

\subsection{Conformal Prediction}
\label{apdx:disc:cp}

%% intro
Conformal prediction \cite{vovk2005algorithmic} provides a promising way to quantify uncertainty of a model
with a correctness guarantee under minimal assumptions.
Here, we consider PAC prediction sets \cite{Park2020PAC},
an interpretation of 
tolerance region \cite{wilks1941determination} and training-conditional inductive conformal prediction \cite{vovk2013conditional}
in the lens of PAC learning theory \cite{valiant1984theory}
(\ie learning a ``good'' function within a function family from data).
This interpretation inspires us to generalize selective generation for GLMs via neural selection functions.
% In conformal prediction, we let
% $\Xs$ be an example space,
% $\Ys$ be a label space,
% and
% $D$ be a distribution over $\Xs \times \Ys$.

%% {\color{blue}
%% %% prediction set model / conformal predictor
%% \para{{\color{red}Conformal Set} Models.}
%% Let $\Xs$ be example space,
%% $\Ys$ be label space,
%% and
%% $D$ be a distribution over $\Xs \times \Ys$.
%% We consider a \emph{{\color{red}conformal (prediction) set model}} $\Ch: \Xs \to 2^{\Ys}$ with a scalar parameter \cite{papadopoulos2002inductive,Park2020PAC} as follows:
%% \begin{align}
%%   C(x) \coloneqq \left\{ y \in \Ys \mid f(x, y) \ge \tau \right\},
%%   \label{eq:ps}
%% \end{align}
%% where
%% $f: \Xs \times \Ys \to \realnum_{\ge 0}$ is a \emph{scoring function}
%% that measures the conformity (or likelihood) of $x$ for being $y$ with respect to $f$ (thus, we call $f(x, y)$ a \emph{conformity score}),
%% and
%% $\tau \in \realnum_{\ge 0}$ is a scalar parameter.
%% We denote the family of all prediction sets with a scalar parameter by $\Ts$.
%% }

\para{Conformal Set Model.}
% conformal set model
We consider a \emph{conformal (prediction) set model} $\Ch: \Xs \to 2^{\Ys}$
that measures the uncertainty of a target model; in conformal prediction,
this model is specifically called a \emph{scoring function}
$f: \Xs \times \Ys \to \realnum_{\ge 0}$ 
that measures the conformity (or likelihood) of $\x$ for being $\y$ with respect to $f$;
thus, $f(\x, \y)$ is called a \emph{conformity score}.
In particular, we consider scalar parameterization of a conformal set \cite{Park2020PAC,papadopoulos2002inductive} as follows:
$
%\begin{equation}
  C(\x) \coloneqq \left\{ \y \in \Ys \mid f(\x, \y) \ge \tau \right\},
  %\label{eq:ps}
%\end{equation}
$
where
$\tau \in \realnum_{\ge 0}$ is a scalar parameter.

% uncertainty
\para{Conformal Sets and Uncertainty.}
% conformal set
The output of the conformal set model is a set of labels,
which naturally represents the \emph{uncertainty of a scoring function on an example} via the size of a conformal set.
In particular, if the scoring function $f$ is unsure on its prediction on $\x$
(due to uncertainty on a label distribution of $\x$, \ie aleatoric uncertainty, and
due to uncertainty in the modeling of $f$, \ie epistemic uncertainty),
the conformal set is larger than it is when the scoring function is sure on its prediction.

To be precise, we consider a \emph{true conformal set} $C^*(\x) \coloneq \{ \y \in \Ys \mid f(\x, \y) \ge f(\x, \y^*)\}$, where $\y^*$ is the true label of $x$.
In particular, the true conformal set is a minimal set that contains a true label and labels with larger scores than the true label score; thus, the size of the true conformal set intuitively measures the uncertainty of a scoring function on the given example, \ie the scoring function's possibilities on making wrong predictions, instead of the true prediction.

% approximation
The true conformal set clearly captures the uncertainty, but the true label is unknown in inference time. Thus, the true conformal set is approximated via scalar parameterization \cite{Park2020PAC,papadopoulos2002inductive} as follows:
\begin{equation}
  C(\x) \coloneqq \left\{ \y \in \Ys \mid f(\x, \y) \ge \tau \right\},
  \label{eq:ps}
\end{equation}
where
$\tau \in \realnum_{\ge 0}$ is a scalar parameter.
%We denote the family of all conformal sets with a scalar parameter by $\Ts$.

% correctness
\para{Correctness.}
As we desire to construct a conformal set close to the true conformal set,
we define the correctness of the conformal set based on its similarity to the true one.
In particular, we wish to have the smallest $C(\x)$ such that $C^*(\x) \subseteq C(\x)$,
or equivalently $C(\x)$ needs to have the smallest $\tau$ while $y \in C(\x)$.
This correctness definition is realized into two ways: a coverage guarantee \cite{vovk2005algorithmic} or a PAC guarantee \cite{vovk2013conditional}.
%In this paper, we specifically consider the PAC guarantee, but our method holds also with the coverage guarantee.
%In the following, we consider the definition of the PAC guarantee along with a necessary assumption.

%% assumption
\para{Assumption.}
We assume that samples are independent and identically distributed (i.i.d.), \ie the i.i.d. assumption. In particular, all samples for testing and learning prediction sets are independently drawn from the same but known distribution $\Ds$.

%% error
\para{PAC guarantee.}
Under the i.i.d. assumption,
we learn a conformal set $\Ch$ that includes the most true labels (\emph{approximately correct}).
In particular,
this means that
the miscoverage of $\Ch$ is less than a desired level $\epsilon \in (0, 1)$, \ie
$
  %\label{eq:miscoverage}
  \Rs_{\text{MC}}(\Ch) \coloneqq \Prob\{ \y \notin \Ch(\x) \} \le \epsilon, 
$
where the probability is taken over i.i.d. samples $(\x, \y) \sim \Ds$.
This risk on micoverage can be generalized to be the risk on indicator loss, $\Rs_{01}(\Ch) \coloneqq \Exp_{\Ds} \ell_{01}(\Ch, \x, \y)$. 
Here, 
the conformal set $\Ch$ is learned from a randomly drawn calibration set, so
we desire to construct $\Ch$ that has a desired error for the most of random calibration sets (\emph{probably approximately correct}), \ie
$
%\begin{align}
  \Prob\{ \Rs_{01}(\Ch) \le \epsilon \} \ge 1 - \delta, 
  %\label{eq:pac}
%\end{align}
$
where $\delta \in (0, 1)$ is a desired confidence level and
the probability is taken over $n$ i.i.d. calibration samples $\Z \sim \Ds^n$,
used to learn $\Ch$.

%% algorithm
\para{Algorithm.}
The PAC conformal prediction set method \cite{Park2020PAC,Park2022PAC} considers the following algorithm $\As_{\text{Binom}}: (\Xs \times \Ys)^* \to \Hs$ to learn a conformal set model $\Ch$, parameterized by $\hat\tau$, where $\Hs$ is a finely-discretized $\realnum_{\ge0}$:
\begin{equation}
  \As_{\text{Binom}}\footnote{$\As_{\text{Binom}}$ returns $\hat\tau = 0$ if it is infeasible.}:
  \quad \hat\tau = \max_{\tau \in \Hs}~\tau \quad \text{subj. to}\quad
  U_{\text{Binom}}(k_\tau; n, \delta) \le \epsilon, \label{eq:pacpsalg}
\end{equation}
where
$k_\tau \coloneqq \sum_{i=1}^n \ell_{01}(\Ch, \x_i, \y_i)$.
Here, $U_\text{Binom}$ is a binomial tail bound, \ie $\Prob\left\{ \Rs_{01}(C) \le U_{\text{Binom}}(k_\tau; n, \delta) \right\} \!\ge\! 1 - \delta$ for any $C$,
where
$U_{\text{Binom}}(k; n, \delta) \!\coloneqq\! \inf \left\{ \theta \in [0, 1] \!\mid\! F(k; n, \theta) \!\le\! \delta \right\} \!\cup\! \{1\}$
and
$F(k; n, \theta)$ is a cumulative distribution function (CDF) of a binomial distribution with $n$ trials and success probability $\theta$.
%% $C$ is parameterized by $\tau$,
%% $\bar{L}_n(C) \!\coloneqq\! \sum_{i=1}^n \! \mathbbm{1}\( y_i \!\notin\! C(x_i) \)$ is the empirical error count of $C$,
%% $F(k; n, \theta)$ is a cumulative distribution function of a binomial distribution with $n$ trials and success probability $\theta$,
%% and
%% $U_{\text{Binom}}(C, Z, \delta) \!\coloneqq\! \inf \left\{ \theta \in [0, 1] \!\mid\! F(\bar{L}_n(C); n, \theta) \!\le\! \delta \right\} \!\cup\! \{1\}$
%% is a binomial tail bound, \ie $\Prob\!\left\{ L(C) \!\le\! U_{\text{Binom}}(C, Z, \delta) \right\} \!\ge\! 1 - \delta$ for any $C$.
%% pac algorithm and guarantee
This algorithm is PAC.
\begin{theorem}{(\cite{Park2020PAC,vovk2013conditional,Park2022PAC})}
  \label{thm:pacps}
  The algorithm $\As_{\text{Binom}}$ is PAC, \ie
  for any $f$, $\epsilon \in (0, 1)$, $\delta \in (0, 1)$, and $n \in \integernum_{\ge 0}$,
  we have
  $\Prob\{ \Rs_{01}(\Ch) \le \epsilon \} \ge 1 - \delta$,
  where
  %the inner probability is taken over a labeled example $(\x, \y) \sim \D$,
  the probability is taken over i.i.d. labeled examples $\Z \sim \Ds^n$,
  and
  $\Ch = \As_{\text{Binom}}(\Z)$.
\end{theorem}
Here, we slightly generalize the known PAC guarantee to hold for any risk with indicator loss. See \autoref{apdx:proof:thm1} for a proof. 
% note on the sample complexity
Note that the PAC guarantee generally holds only if an enough number of samples is provided
(when we know a function family including a true function).
However, we consider PAC algorithms that hold for any number of samples due to the structural property of prediction sets, \ie a prediction set is always correct if $\tau = 0$ (thus $\smash\Ch(\x) = \Ys$), regardless of the sample size.
In other words, if the calibration samples are not sufficient, the prediction set is constructed to return $\Ys$ to satisfy the PAC guarantee.

\subsection{Sample Space Decomposition}
Given the generator $G$ and the entailment set function $\hat{E}$, the sample space $\Omega \coloneqq \mathcal{X} \times \mathcal{Y} \times \Es \times \Vs$ can be partitioned as follows: 
\begin{align*}
    \Omega&=\underbrace{\{(\x,\y,e, v) \ \vert \  G(\x)\in E_{\text{true}}(\y)\}}_{\Omega_{\text{TD}}^{E_\text{true}}}\cup\underbrace{\{(\x,\y,e, v) \ \vert \ G(\x)\notin E_{\text{true}}(\y)\}}_{\Omega_{\text{FD}}^{E_\text{true}}}\\            
    &=\underbrace{\{(\x,\y,e, v) \ \vert \  e=0\}}_{\Omega_{\text{TD}}^{E_\text{true}}}\cup\underbrace{\{(\x,\y,e, v) \ \vert \ e=1\}}_{\Omega_{\text{FD}}^{E_\text{true}}}\\           
    &=\underbrace{\underbrace{\{(\x,\y,e, v) \ \vert \  e=1 \text{ and } G(\x)\in\hat{E}(\y)\}}_{\Omega_{\text{TE}}^{\hat{E}}}\cup\underbrace{\{(\x,\y,e, v) \ \vert \ e=1\text{ and } G(\x)\notin\hat{E}(\y)\}}_{\Omega_{\text{FNE}}^{\hat{E}}}}_{\Omega_\text{TD}}\cup\\
    &\quad\quad\quad\quad\quad\underbrace{\underbrace{\{(\x,\y,e, v) \ \vert \  e=0 \text{ and } G(\x)\notin\hat{E}(\y)\}}_{\Omega_{\text{TNE}}^{\hat{E}}}\cup\underbrace{\{(\x,\y,e, v) \ \vert \ e=0 \text{ and } G(\x)\in\hat{E}(\y)\}}_{\Omega_{\text{FE}}^{\hat{E}}}}_{\Omega_\text{FD}}\\             &=\underbrace{\big\{ \Omega_{\text{TE}}^{\hat{E}}\cup\Omega_{\text{FE}}^{\hat{E}} \big\}}_{\Omega_{\text{TD}}^{\hat{E}}} \cup \underbrace{\big\{\Omega_{\text{FNE}}^{\hat{E}}\cup\Omega_{\text{TNE}}^{\hat{E}}\big\}}_{\Omega_{\text{FD}}^{\hat{E}}}.
\end{align*}

Here, the short-hands are defined as follows:
\begin{itemize}
[leftmargin=*,topsep=0pt,itemsep=0ex,partopsep=1ex,parsep=0ex]
    \item True discovery rate (TDR): $\mathbb{P}(\Omega_{\text{TD}}^{E_\text{true}})$
    \item False discovery rate (FDR): $\mathbb{P}(\Omega_{\text{FD}}^{E_\text{true}})$
    \item True entailment rate (TER): $\mathbb{P}(\Omega_{\text{TE}}^{\hat{E}})$
    \item False non-entailment rate (FNER): $\mathbb{P}(\Omega_{\text{FNE}}^{\hat{E}})$
    \item True non-entailment rate (TNER): $\mathbb{P}(\Omega_{\text{TNE}}^{\hat{E}})$
    \item False entailment rate (FER): $\mathbb{P}(\Omega_{\text{FER}}^{\hat{E}})$
\end{itemize}

\subsection{Experiment Setup}
\label{sec:apdx:expsetup}

\subsubsection{Computing Environment}
Our system environment consists of 4 NVIDIA A100 80GB with 128 CPUs.

\subsubsection{Models and Datasets} 
\label{sec:apdx:modelsdatasets}

We use two large language models (LLMs), \textit{GPT-3.5-Turbo} and \textit{Alpaca-7B}, for language generation. 
We use deberta-v2-xxlarge-mnli as our entailment model.

For each GLM to annotate entailment labels for each question, answer, and generated answer pair, we annotate entailment labels. 
Specifically, we consider the open-ended QA task, where the model is prompted to generate the answer in a declarative form given a question. To validate our method and its theoretical guarantee on controlling FDR-E, we create a dataset on textual entailment using the Natural Questions (NQ) dataset \cite{kwiatkowski2019natural} for each GLM. 
Based on the transformation method by \cite{demszky2018transforming} that converts the question and answer pair in QA dataset into a declarative form, we manually labeled textual entailment by letting the generated sequence as the premise and the reference answer in declarative form as the hypothesis. 
Similar work can be found in \cite{chen_can_2021}, but they label the textual entailment based on the extractive answer from the model.
Approximately 7.3k (7,374) and 4.6k (4,595) samples are labeled for \textit{Alpaca-7B} and \textit{GPT-3.5-Turbo}, respectively, and both are split into calibration and test data at an 8:2 ratio. 
For semi-supervised learning algorithms that exploit unlabeled data (\autoref{alg:nsegen_additional_delta}, \autoref{alg:selgenpsl}), at most 27k and 10k unlabeled samples are used to train a selective generator, varying its size. 
Besides, semi-supervised learning algorithms use only 75\% of the labeled calibration data compared to what is used by supervised methods (\autoref{alg:selgenel}, \autoref{alg:selgenem}).

\section{Semi-supervised Selective Generation Algorithms (Certified)} \label{sec:sel_ent_algs}

%%%%%%%%%%%%%%%%%%%%%%%%%%%%%%%%%%%%%%%%%%%%%%%%%%%%%%%%%%%%%%%%%%%%%%%%%%%%%%%%%%%%%%%%%%%%%%%%%

% algorithm: entailment set learning
\begin{algorithm}[ht]
    \caption{ Entailment Set Learning with a False Entailment Rate (FER) Guarantee }
    \label{alg:es}

    \begin{algorithmic}[1]
    \Procedure{ES}{$f_E$, $\Z_E$, $\epsilon_E$, $\delta_E$}

    \State $\Z_E \gets \textsc{Sort}_{f_E}(\Z_E)$
    $\Comment{\text{In an increasing order of $f_E(\y_i, G(\x_i))$ }}$
    \State $(\underline{i}, \overline{i}) \gets (1, |\Z_E|)$

    \For{$i=1$ {\bfseries to } $\lceil \text{log} | \Z_E | \rceil$}    
        
        \State $k^{(i)} \gets \sum_{(\x, \y, e) \in \Z_E} \mathbbm{1}(e=0, f_E(G(\x), \y) \geq f_E(G(\x_{\lceil (\underline{i} + \overline{i})/2 \rceil}), \y_{\lceil (\underline{i} + \overline{i})/2 \rceil}))$
        \State $U \gets U_{\text{Binom}}( k^{(i)}, | \Z_E |, \delta_E )$

        \If{$U \leq \epsilon_E$}
            \State $\overline{i} \gets \lceil (\underline{i} + \overline{i})/2 \rceil$
        \Else 
            \State $\underline{i} \gets \lceil (\underline{i} + \overline{i})/2 \rceil$
        \EndIf

    \EndFor

    \State {\bfseries return} $\tau_E$
    \EndProcedure

    \end{algorithmic}
    
\end{algorithm}

%%%%%%%%%%%%%%%%%%%%%%%%%%%%%%%%%%%%%%%%%%%%%%%%%%%%%%%%%%%%%%%%%%%%%%%%%%%%%%%%%%%%%%%%%%%%%%%%%

% algorithm: U_SSL computation
\begin{algorithm}[ht]
    \caption{ $U_{\text{SSL}}$ Computation (for Single $\epsilon_E$) }
    \label{alg:u_ssl}

    \begin{algorithmic}[1]
    \Procedure{Compute-$U_{\text{SSL}}$}{$f_E$, $\Z_E$, $\Z_U$, $\delta_S$, $\epsilon_E$, $\delta_E$}

    \State $\tau_E \gets \texttt{ES}( f_E, \Z_E, \epsilon_E, \delta_E/2 )$

    \State $\ell \gets \sum_{(\x, \y, e) \in \Z_E} \mathbbm{1} ( e=1, f_E( G(\x), \y ) < \tau_E )$

    \State $k \gets \sum_{(\x, \y) \in \Z_U} \mathbbm{1} (f_E( G(\x), \y) < \tau_E ) $

    \State $U_{\text{SSL}} \gets 
        \epsilon_E 
        - L_{\text{Binom}}( \ell; | \Z_E |, \delta_E/2 ) 
        + U_{\text{Binom}}( k, | \Z_U |, \delta_S/2 )$

    % \State {\bfseries return} $k$, $U_{\text{SSL}}$
    \State {\bfseries return} $U_{\text{SSL}}$
    \EndProcedure

    \end{algorithmic}
    
\end{algorithm}

%%%%%%%%%%%%%%%%%%%%%%%%%%%%%%%%%%%%%%%%%%%%%%%%%%%%%%%%%%%%%%%%%%%%%%%%%%%%%%%%%%%%%%%%%%%%%%%%%

% algorithm: Optimal U_SSL search
\begin{algorithm}[ht]
    \caption{ Optimal $U_{\text{SSL}}$ Search }
    \label{alg:opt_u_ssl}

    \begin{algorithmic}[1]
    \Procedure{Compute-$U_{\text{SSL}}^{\text{opt}}$}{$f_E$, $\Z_E$, $\Z_U$, $\delta_S$, $Q$, $\delta_E$}

    \State $\Z_E \gets \textsc{Sort}_{f_E}(\Z_E)$
    $\Comment{\text{In an increasing order of $f_E(\y_i, G(\x_i))$}}$
    \State $(\underline{i}, \overline{i}) \gets (1, |\Z_E|)$

    \State 
    $
    \epsilon_{\text{max}} 
    \gets
    \sum_{(\x, \y, e) \in \Z_E} \mathbbm{1}(e=0) / | \Z_E |
    $

    \State 
    $\mathcal{H}_E \gets \{ \epsilon_1 = \epsilon_{\text{max}}, \dots, \epsilon_Q = 1/|Q| \epsilon_{\text{max}} \}$

    \State 
    $U_{\text{SSL}}^{\text{OPT}} \gets \infty$

    \For{$i$ {\bfseries in } $\{1, \dots, Q\}$}    
        % \State 
        % \textcolor{red}{$k^{(i)}, U_{\text{SSL}}^{(i)} 
        % = \texttt{Compute-$U_{\text{SSL}}$}( f_E, \Z_E, \Z_U, \delta_S/Q, \epsilon_i, \delta_E/Q )$}
        \State 
        $U_{\text{SSL}}^{(i)} 
        \gets \texttt{Compute-$U_{\text{SSL}}$}( f_E, \Z_E, \Z_U, \delta_S/Q, \epsilon_i, \delta_E/Q )$
        
        \If{$U_{\text{SSL}}^{(i)} \leq U_{\text{SSL}}^{\text{OPT}}$}
        % \State \textcolor{red}{$\tau_E \gets \tau$} \SP{remove this?}
        \State $U_{\text{SSL}}^{\text{OPT}} \gets U_{\text{SSL}}^{(i)}$
        \EndIf
    \EndFor

    \State {\bfseries return} $U_{\text{SSL}}^{\text{OPT}}$
    \EndProcedure

    \end{algorithmic}
    
\end{algorithm}

\begin{algorithm}[ht]
    \caption{FDR-E Bound Computation}
    \label{alg:fdrbound_additional_delta}
    
    \begin{algorithmic}[1]
    
    %% ---------- upper bound
    \Procedure{FDR-E-Bound}{$f_E$, $\Z_E$, $\Z_U$, $\delta_S$, $Q$, $\delta_E$, $\delta_W$}

    \State $w_\text{SL} \gets U_{\text{Binom}}(|\Z_E|; |\Z_E|+|\Z_U|, \delta_W / 2 )$
    $\Comment{\text{Upper bound of (B) in (\ref{eq:fdrdecomp-ss-ssl})}}$
    
    \State $k_\text{SL} \gets \sum_{(\x, \y, e) \in \Z_E} \mathbbm{1}( e=0 )$
    \State $U_{\text{SL}} \gets U_{\text{Binom}}(k_\text{SL}; |\Z_E|, \delta_S / 2 )$ %\label{alg:unionbnd}
    $\Comment{\text{Upper bound of (C) in (\ref{eq:fdrdecomp-ss-ssl})}}$
    
    % \State $l \gets \sum_{(\x, \y, e) \in \Z_E} \mathbbm{1}(e=1, f_E(G(\x), \y) < \tau_E)$ \SP{rmove this?}
    % \State $k \gets \sum_{(\x, \y) \in \Z_U} \mathbbm{1}(f_E(G(\x), \y) < \tau_E)$  \SP{rmove this?}

    \State $w_\text{SSL} \gets U_{\text{Binom}}(|\Z_U|; |\Z_E|+|\Z_U|, \delta_W / 2 )$
    $\Comment{\text{Upper bound of (D) in (\ref{eq:fdrdecomp-ss-ssl})}}$
    
    \State $U_{\text{SSL}}^{\text{OPT}} \gets \texttt{Compute$-U_{\text{SSL}}^{\text{OPT}}$}(f_E, \Z_E, \Z_U, \delta_S/2, Q, \delta_E/2)$
    $\Comment{\text{Upper bound of (E) in (\ref{eq:fdrdecomp-ss-ssl})}}$
    
    \State $U \gets w_{\text{SL}} U_{\text{SL}} + w_{\text{SSL}} U_{\text{SSL}}^{\text{OPT}}$
    \State {\bfseries return} $U$ 
    \EndProcedure
    \end{algorithmic}
\end{algorithm}

%%%%%%%%%%%%%%%%%%%%%%%%%%%%%%%%%%%%%%%%%%%%%%%%%%%%%%%%%%%%%%%%%%%%%%%%%%%%%%%%%%%%%%%%%%%%%%%%%

% algorithm: a single parameter
\begin{algorithm}[ht]
    \caption{Semi-supervised Selective Generator Learning (Single-threshold Selection Function)}
    \label{alg:sgsingle_additional_delta}

    \begin{algorithmic}[1]
    \Procedure{SGen-Semi}{$f_M$, $f_E$, $G$, $\Z_{E}$, $\Z_U$, $\epsilon_S$, $\delta_S$, $Q$, $\delta_E$, $\delta_W$, $\texttt{return\_bool}=\texttt{False}$}

    \State {$\Z_{U,E} \gets \Z_U \cup \Z_E$}
    \State $\Z_{U, E} \gets \textsc{Sort}_{f_M}(\Z_{U, E})$
    $\Comment{\text{In an increasing order of $f_M(\y_i, G(\x_i))$}}$
    
    \State $(\underline{i}, \overline{i}) \gets (1, \Z_{U, E})$
    \State $U_\text{min} \gets \infty; \tau_{\text{min}} \gets \texttt{NULL}$
    % \State $\texttt{TER}_\text{max} \gets 0$
    \For{$i=1$ {\bfseries to} $\lceil \log_2 \Z_{U, E} \rceil$}
    
    \State $\tau_S^{(i)} \gets f_M(\x_{\lceil{(\underline{i} + \overline{i})}/{2}\rceil}, G(\x_{\lceil{(\underline{i} + \overline{i})}/{2}\rceil}))$
    \State $\Z_E^{(i)} \gets \{ (\x, \y, e) \in \Z_E \mid f_M(\x, G(\x)) \ge \tau_S^{(i)}  \}$
    \State $\Z_U^{(i)} \gets \{ (\x, \y) \in \Z_U \mid f_M(\x, G(\x)) \ge \tau_S^{(i)} \}$
    
    \State $U^{(i)} \gets \textsc{FDR-E-Bound}(f_E, \Z_E^{(i)}, \Z_U^{(i)}, \frac{\delta_S}{\lceil \log_2 |\Z_{U, E}| \rceil}, Q, \frac{\delta_E}{\lceil \log_2 \Z_{U, E} \rceil}, \frac{\delta_W}{\lceil \log_2 \Z_{U, E} \rceil} )$

    \If{$U^{(i)} \leq U_{\text{min}}$}
        \State $U_{\text{min}} \gets U^{(i)} ;~ \tau_{\text{min}} \gets \tau_S^{(i)}$
    \EndIf

    % \State $U_{\text{min}} \gets \text{min}(U_{\text{min}}, U^{(i)})$
    \If{$U^{(i)} \le \ep_S$} 
    \State $\overline{i} \gets \lceil (\underline{i} + \overline{i})/2 \rceil$
    \Else
    \State $\underline{i} \gets \lceil (\underline{i} + \overline{i})/2 \rceil$ 
    \EndIf
    \EndFor
    \State $\tau_S \gets \tau_S^{(i)}$

    \If{$U_{\min} \le \ep_S$} 
        \State $\hat U \gets U^{(i)}$ 
        \State $\texttt{Bounded} \gets \texttt{Success}$
    \Else
    \State $\hat U \gets U_{\min}$
    \State $\tau_S \gets \tau_{\text{min}}$
    \State ${\texttt{Bounded} \gets \texttt{Fail}}$ 
    \EndIf    
    \State {\bfseries return} $(\tau_S, \hat U, \texttt{Bounded})$ if $\texttt{return\_bool}$ else $(\tau_S, \hat U)$.
    \EndProcedure
    \end{algorithmic}
\end{algorithm}

%%%%%%%%%%%%%%%%%%%%%%%%%%%%%%%%%%%%%%%%%%%%%%%%%%%%%%%%%%%%%%%%%%%%%%%%%%%%%%%%%%%%%%%%%%%%%%%%%

% algorithm: two-dimensional parameter
\begin{algorithm}[ht]
    \caption{Semi-supervised Selective Generator Learning (Double-threshold Selection Function)}
    \label{alg:sgdouble_additional_delta}

    \begin{algorithmic}[1]
    \Procedure{SGen-Semi2}{$f_{M_1}$, $f_{M_2}$, $f_E$, $G$, $\Z_E$, $\Z_U$, $\epsilon_S$, $\delta_S$, $Q$, $\delta_E$, $\delta_W$, $\texttt{return\_bool}=\texttt{False}$}
    \State $\Z_{U,E} \gets \Z_U \cup \Z_E$
    \State $\Z_{U_1, E_1} \gets \textsc{Sort}_{f_{M_1}}(\Z_{U, E})$
    $\Comment{\text{In an increasing order of $f_{M_1}(\y_i, G(\x_i))$}}$
    \State $\Z_{U_2, E_2} \gets \textsc{Sort}_{f_{M_2}}(\Z_{U, E})$
    $\Comment{\text{In an increasing order of $f_{M_2}(\y_i, G(\x_i))$}}$
    
    \State $U_\text{min} \gets \infty ;~ \tau_{\text{min}} \gets \texttt{NULL}$
    \State $(\underline{i}, \overline{i}) \gets (1, |\Z_{U_1, E_1}|)$
    \State $I \gets \lceil \log_2 |\Z_{U, E}| \rceil$

    % first iteration
    \For{$i=1$ {\bfseries to} $\lceil \log_2 |\Z_{U, E}| \rceil$}
    
    \State $\tau_S^{(i)} \gets f_{M_1}(\x_{\lceil{(\underline{i} + \overline{i})}/{2}\rceil}, G(\x_{\lceil{(\underline{i} + \overline{i})}/{2}\rceil}))$

    % second iteration
    \State $U_\text{min}^{(i)} \gets \infty ;~ \tau_{\text{min}}^{(i)} \gets \texttt{NULL}$
    \State $(\underline{j}, \overline{j}) \gets (1, |\Z_{U_2, E_2}|)$
    \For{$j=1$ {\bfseries to} $\lceil \log_2 |\Z_{U, E}| \rceil$}
    \State $\tau_S^{(j)} \gets f_{M_2}(\x_{\lceil{(\underline{j} + \overline{j})}/{2}\rceil}, G(\x_{\lceil{(\underline{j} + \overline{j})}/{2}\rceil}))$
    
    \State $\Z_E^{(i,j)} \gets \{ (\x, \y, e) \in \Z_E \mid \sh(\x; G, f_{M_1}, f_{M_2}, \tau_S^{(i)}, \tau_S^{(j)}) = 1 \}$
    \State $\Z_U^{(i,j)} \gets \{ (\x, \y) \in \Z_U \mid \sh(\x; G, f_{M_1}, f_{M_2}, \tau_S^{(i)}, \tau_S^{(j)}) = 1 \}$
    \State $U^{(i,j)} \gets \textsc{FDR-E-Bound}(f_E, \Z_E^{(i,j)}, \Z_U^{(i,j)}, \frac{\delta_S}{I^2}, Q, \frac{\delta_E}{I^2}, \frac{\delta_W}{I^2})$
    
    \If{$U^{(i, j)} \leq U_{\text{min}}^{(i)}$}
        \State 
        $U_{\text{min}}^{(i)} \gets U^{(i,j)}; \tau_{\text{min}}^{(i)} \gets (\tau_S^{(i)}, \tau_S^{(j)})$
    \EndIf
    
    \If{$U^{(i,j)} \le \ep_S$} 
        \State $\overline{j} \gets \lceil (\underline{j} + \overline{j})/2 \rceil$
    \Else
        \State $\underline{j} \gets \lceil (\underline{j} + \overline{j})/2 \rceil$ 
    \EndIf
    \EndFor

    \If{$U_{\text{min}}^{(i)} \leq U_{\text{min}}$}
        \State $U_{\text{min}} \gets U_{\text{min}}^{(i)} ;~ \tau_{\text{min}} \gets \tau_{\text{min}}^{(i)}$
    \EndIf

    \If{$i \neq \lceil \text{log}_2 | \Z_{U, E} | \rceil$}
        \If{$U_{\text{min}}^{(i)} \le \ep_S$} 
        % \State $\texttt{TER}_{\text{max}} \gets \texttt{TER}$
            \State $\overline{i} \gets \lceil (\underline{i} + \overline{i})/2 \rceil$
        \Else
            \State $\underline{i} \gets \lceil (\underline{i} + \overline{i})/2 \rceil$ 
        \EndIf    
    \Else
        \State $\tau_S \gets (\tau_S^{(i)}, \tau_S^{(j)})$
    \EndIf
    
    \EndFor

    \If{$U_{\min} \le \ep_S$} 
        \State $\hat U \gets U^{(i,j)} ;~ \texttt{Bounded} \gets \texttt{Success}$
    \Else
        \State $\hat U \gets U_{\min} ;~ \tau_S \gets \tau_{\text{min}} ;~ \texttt{Bounded} \gets \texttt{Fail}$ 
    \EndIf    
    %\State $(\tau, U_{\text{end}}) \gets (f(\x_{i}, \y_{i}), U_{\text{min}})$
    \State {\bfseries return} $(\tau_S, \hat U, \texttt{Bounded})$ if $\texttt{return\_bool}$ else $(\tau_S, \hat{U})$ 
    \EndProcedure
    \end{algorithmic}
\end{algorithm}

\begin{algorithm}[ht!]
    \caption{Semi-supervised Selective Generator Learning with Neuro-Selection}
    \label{alg:nsegen_additional_delta}
    \begin{algorithmic}[1]

    \Procedure{SGen-Semi-MS}{$f_{M_1}$, $f_{M_2}$, $f_E$, $G$, $\Z_E$, $\Z_U$, $\epsilon_S$, $\delta_S$, $Q$, $\delta_E, \delta_W$}

        \State $\mathcal{M}_{\text{Success}}=\{ \} ;~ \mathcal{M}_{\text{Fail}}=\{ \}$

        \State \scalebox{0.75}{$(\tau_{S_1}, \hat{U}_1, \texttt{Bounded}_1) \gets \texttt{SGen-Semi}$($f_{M_1}$, $f_E$, $G$, $\Z_{E}$, $\Z_U$, $\epsilon_S$, $\delta_S/3$, $Q$, $\delta_E/3$, $\delta_W/3$, $\texttt{return\_bool}=\texttt{True}$)} 

        \State \scalebox{0.75}{$(\tau_{S_2}, \hat{U}_2, \texttt{Bounded}_2) \gets \texttt{SGen-Semi}$($f_{M_2}$, $f_E$, $G$, $\Z_{E}$, $\Z_U$, $\epsilon_S$, $\delta_S/3$, $Q$, $\delta_E/3$, $\delta_W/3$, $\texttt{return\_bool}=\texttt{True}$)}

        \State \scalebox{0.75}{$(\tau_{S_3}, \hat{U}_3, \texttt{Bounded}_3) \gets \texttt{SGen-Semi2}$($f_{M_1}$, $f_{M_2}$, $f_E$, $G$, $\Z_{E}$, $\Z_U$, $\epsilon_S$, $\delta_S/3$, $Q$, $\delta_E/3$, $\delta_W/3$, $\texttt{return\_bool}=\texttt{True}$)}

        \State \scalebox{0.9}{$\mathcal{M} \coloneqq \{ 
            (\tau_{S_1}, \hat{U}_1, s_1, \texttt{Bounded}_1),
            (\tau_{S_2}, \hat{U}_2, s_2, \texttt{Bounded}_2),
            (\tau_{S_3}, \hat{U}_3, s_3, \texttt{Bounded}_3)
        \} $}

        \State \hfill $\Comment{s_i\text{ refers to the scoring function(s) used in each algorithm.}}$

        %\For{$(\tau_{S}, \Uh, \texttt{Bounded}))$ {\bfseries in} $\{1, 2, 3\}$}
        \For{$(\tau_{S}, \Uh, s, \texttt{Bounded})$ {\bfseries in} $\Ms$}
            \If{$\texttt{Bounded}=\texttt{Success}$}
                \State  $\mathcal{M}_{\text{Success}} \gets \mathcal{M}_{\text{Success}} \cup \{(\tau_{S}, \hat{U}, s)\}$
            \Else
                \State $\mathcal{M}_{\text{Fail}} \gets \mathcal{M}_{\text{Fail}} \cup \{(\tau_{S}, \hat{U}, s)\}$
            \EndIf
        \EndFor

        \If{$\mathcal{M}_{\text{Success}}=\{ \}$}
            \State {\bfseries return} $(\tau_S, \hat{U}, s) \gets \arg\min_{(\tau_{S}, \hat{U}, s) \in \mathcal{M}_{\text{Fail}}}~\hat{U}$
        \Else
            \State {\bfseries return} $(\tau_S, \hat{U}, s) \gets \arg\max_{(\tau_{S}, \hat{U}, s) \in \mathcal{M}_{\text{Success}}}~\hat{U}$
        \EndIf

    \EndProcedure
    \end{algorithmic}
\end{algorithm}

\clearpage

\section{Supervised Selective Generation Algorithms (Certified)} \label{sec:sel_gen_algs}

% algorithm
\begin{algorithm}[ht]
    \caption{Supervised Selective Generator Learning with $\mathcal{R}_{R_E}(\hat{S})$ Control } 
    \label{alg:selgenel}
    
    \begin{algorithmic}[1]
        \Procedure{SG-Sup}{$f_{M}$, $G$, $\Z_E$, $\epsilon$, $\delta$}  
        %\State {\bfseries input:} $\Cwh$, $f$, and $((\x_1, \y_1), \dots, (\x_n, \y_n))$, sorted labeled examples of $\Z$ in the increasing order of $f(\x_j, \y_j)$
        \State $\Z_{E} \gets \textsc{Sort}_{f_M}(\Z_{E})$
        
        \State $(\underline{i}, \overline{i}) \gets (1, |\Z_E|)$
        % \State $U_\text{min} \gets \infty$
        \For{$i=1$ {\bfseries to} $\lceil \log_2 |\Z_E| \rceil$}
        
        \State $\tau_S^{(i)} \gets f_M(\x_{\lceil{(\underline{i} + \overline{i})}/{2}\rceil}, G(\x_{\lceil{(\underline{i} + \overline{i})}/{2}\rceil}))$
        \State $\Z_E^{(i)} \gets \{ (\x, \y, e) \in \Z_E \mid f_M(\x, G(\x)) \ge \tau_S^{(i)} \}$
        
        \State $k^{(i)} \gets \sum_{(\x, \y, e) \in \Z_E} \mathbbm{1}( e=0 )$
        
        %\State $i \gets \lceil{(\underline{i} + \overline{i})}/{2}\rceil$
        %\State $n_j \gets \sum_{k=1}^n \mathbbm{1}( f(\x_k, \Gwh(\x_k)) \ge f(\x_i, \y_i) )$
        %\State $n_j \gets \sum_{k=1}^n \mathbbm{1}( f(\x_k, \Gwh(\x_k)) \ge f(\x_i, \Gwh(\x_i)) )$
        %\State $k_j \gets \sum_{k=1}^n \mathbbm{1}( e_k=0 \text{~and~} f(\x_k, \Gwh(\x_k)) \ge f(\x_i, \Gwh(\x_i)) )$
        
        \State $U^{(i)} \gets U_{\text{Binom}}(k^{(i)}; |\Z_E^{(i)}|, \delta / \lceil \log_2 |\Z_{E}| \rceil )$ %\label{alg:unionbnd}
        % \State $U_{\text{min}} \gets \text{min}(U_{\text{min}}, U^{(i)})$
        \If{$U^{(i)} \le \epsilon$} 
        \State $\overline{i} \gets \lceil (\underline{i} + \overline{i})/2 \rceil$
        \Else
        \State $\underline{i} \gets \lceil (\underline{i} + \overline{i})/2 \rceil$    
        \EndIf
        \EndFor
        \State $\tau_S \gets \tau_S^{(i)}$
        \State $\Uh \gets U^{(i)}$
        \State {\bfseries return} $\tau_S, \Uh$
        \EndProcedure  
    \end{algorithmic}
\end{algorithm}

\newpage

\section{Semi-supervised Selective Generation Algorithms (Heuristic)}

% algorithm
\begin{algorithm}[ht]
    \caption{Semi-supervised Selective Generator Learning with Pseudo-entailment Labels} 
    \label{alg:selgenpsl}
    
    \begin{algorithmic}[1]
        \Procedure{SG-PSL-H-Semi}{$f_M$, $f_E$, $G$, $\Z_E$, $\Z_U$, $\epsilon$, $\delta$, $\tau_{\text{PL}}$, $\texttt{FILTER}$}  
        %\State {\bfseries input:} $\Cwh$, $f$, and $((\x_1, \y_1), \dots, (\x_n, \y_n))$, sorted labeled examples of $\Z$ in the increasing order of $f(\x_j, \y_j)$
        \If{$\texttt{FILTER}==\texttt{TRUE}$}
        \State $\Z_U \leftarrow \{ (\x, \y)~\vert~f_E( G(\x), \y ) \geq \tau_{\text{PL}}~\text{or}~1 - f_E( G(\x), \y ) \geq \tau_{\text{PL}} \}$  
        \EndIf
        \State $\Z_U \leftarrow \{ ( \x, \y, \tilde{e} )~\vert~( \x, \y) \in \Z_U, \tilde{e} = \mathbbm{1}\big( f_E( G(\x), \y ) \geq \tau_{\text{PL}} \big) \}$
        \State $\Z_E \leftarrow \{ ( \x, \y, \tilde{e} )~\vert~( \x, \y, e ) \in \Z_U, \tilde{e} = e \}$
        \State $\Z_{U, E} \gets \textsc{Sort}_{f_M}(\Z_E \cup \Z_U)$
        \State $(\underline{i}, \overline{i}) \gets (1, |\Z_{U, E}|)$
        % \State $U_\text{min} \gets \infty$
        \For{$i=1$ {\bfseries to} $\lceil \log_2 |\Z_{U, E}| \rceil$}
        
        \State $\tau_S^{(i)} \gets f_M(\x_{\lceil{(\underline{i} + \overline{i})}/{2}\rceil}, G(\x_{\lceil{(\underline{i} + \overline{i})}/{2}\rceil}))$
        \State $\Z_{U, E}^{(i)} \gets \{ (\x, \y) \in \Z_{U, E} \mid f_M(\x, G(\x)) \ge \tau_S^{(i)} \}$
        
        \State $k^{(i)} \gets \sum_{(\x, \y, \tilde{e}) \in \Z_{U, E}^{(i)}} \mathbbm{1}( \tilde{e} = 0 )$
        
        %\State $i \gets \lceil{(\underline{i} + \overline{i})}/{2}\rceil$
        %\State $n_j \gets \sum_{k=1}^n \mathbbm{1}( f(\x_k, \Gwh(\x_k)) \ge f(\x_i, \y_i) )$
        %\State $n_j \gets \sum_{k=1}^n \mathbbm{1}( f(\x_k, \Gwh(\x_k)) \ge f(\x_i, \Gwh(\x_i)) )$
        %\State $k_j \gets \sum_{k=1}^n \mathbbm{1}( e_k=0 \text{~and~} f(\x_k, \Gwh(\x_k)) \ge f(\x_i, \Gwh(\x_i)) )$
        
        \State $U^{(i)} \gets U_{\text{Binom}}(k^{(i)}; |\Z_{U, E}^{(i)}|, \delta / \lceil \log_2 |\Z_{U, E}| \rceil )$ %\label{alg:unionbnd}
        % \State $U_{\text{min}} \gets \text{min}(U_{\text{min}}, U^{(i)})$
        \If{$U^{(i)} \le \epsilon$} 
        \State $\overline{i} \gets \lceil (\underline{i} + \overline{i})/2 \rceil$
        \Else
        \State $\underline{i} \gets \lceil (\underline{i} + \overline{i})/2 \rceil$    
        \EndIf
        \EndFor
        \State $\tau_S \gets \tau_S^{(i)}$
        \State $\Uh \gets U^{(i)}$
        \State {\bfseries return} $\tau_S, \Uh$
        \EndProcedure  
    \end{algorithmic}
\end{algorithm}

\section{Unsupervised Selective Generation Algorithms (Certified)} \label{sec:sel_gen_em_alg}

% algorithm
\begin{algorithm}[ht!]
    \caption{Unsupervised Selective Generator Learning with $\mathcal{R}_{\text{EM}}(\hat{S})$ Control \cite{geifman2017selective}} 
    \label{alg:selgenem}
    \begin{algorithmic}[1]
    \Procedure{SG-EM}{$f_{M}$, $G$, $\Z_E$, $\Z_U$, $\epsilon$, $\delta$}
      %\State {\bfseries input:} $\Cwh$, $f$, and $((\x_1, \y_1), \dots, (\x_n, \y_n))$, sorted labeled examples of $\Z$ in the increasing order of $f(\x_j, \y_j)$
      \State {$\Z_{U,E} \gets \Z_U \cup \Z_E$}
      \State $\Z_{U, E} \gets \textsc{Sort}_{f_M}(\Z_{U, E})$
      \State $(\underline{i}, \overline{i}) \gets (1, |\Z_{U,E}|)$
      % \State $U_\text{min} \gets \infty$
      \For{$i=1$ {\bfseries to} $\lceil \log_2 |\Z_{U,E}| \rceil$}
  
      \State $\tau_S^{(i)} \gets f_M(\x_{\lceil{(\underline{i} + \overline{i})}/{2}\rceil}, G(\x_{\lceil{(\underline{i} + \overline{i})}/{2}\rceil}))$
      \State $\Z_{U,E}^{(i)} \gets \{ (\x, \y) \in \Z_{U,E} \mid f_M(\x, G(\x)) \ge \tau_S^{(i)} \}$
      
      %\State $i \gets \lceil{(\underline{i} + \overline{i})}/{2}\rceil$
      %\State $n_j \gets \sum_{k=1}^n \mathbbm{1}( f(\x_k, \Gwh(\x_k)) \ge f(\x_{\lceil{(\underline{i} + \overline{i})}/{2}\rceil}, \Gwh(\x_{\lceil{(\underline{i} + \overline{i})}/{2}\rceil})) )$
      \State $k^{(i)} \gets \sum_{(\x, \y) \in \Z_{U,E}^{(i)}} \mathbbm{1}( G(\x) \neq \y )$
      \State $U^{(i)} \gets U_{\text{Binom}}(k^{(i)}; |\Z_{U,E}^{(i)}|, \delta / \lceil \log_2 |\Z_{U,E}| \rceil )$ %\label{alg:unionbnd}
      % \State $U_{\text{min}} \gets \text{min}(U_{\text{min}}, U^{(i)})$
      \If{$U^{(i)} \le \epsilon$} 
      \State $\overline{i} \gets \lceil (\underline{i} + \overline{i})/2 \rceil$
      \Else
      \State $\underline{i} \gets \lceil (\underline{i} + \overline{i})/2 \rceil$    
      \EndIf
      \EndFor
      \State $\tau_S \gets \tau_S^{(i)}$
    \State $\Uh \gets U^{(i)}$
    \State {\bfseries return} $\tau_S, \Uh$
    \EndProcedure
    \end{algorithmic}
  \end{algorithm}

\clearpage

\section{Proof of \autoref{thm:pacps}}
\label{apdx:proof:thm1}

Let
$C_\tau$ be a prediction set $C$ with a parameter $\tau$,  
$\Hs_{\epsilon} \coloneqq \left\{ \tau \in \Hs \mid \Rs_{01}(C_\tau) > \epsilon \right\}$,
and $\tau^* \coloneqq \inf \Hs_\epsilon$,
where $\Hs$ is finely-discretized non-negative real values.
Then, we have
\begin{align}
    \Prob\Big\{ \Rs_{01}(\As_{\text{Binom}}(\Z)) > \epsilon \Big\} 
    &\le \Prob\Big\{ \exists {\tau \in \Hs_\epsilon}, U_{\text{Binom}}(k_\tau; n, \delta) \le \epsilon \Big\} 
    \nonumber \\
    &\le \Prob\Big\{ U_{\text{Binom}}(k_{\tau^*}; n, \delta) \!\le\! \epsilon \Big\}
    \label{thm:bg:pps:mono}
    \\
    &\le \Prob\Big\{ \Rs_{01}(C_{\tau^*}) > \epsilon \wedge U_{\text{Binom}}(k_{\tau^*}; n, \delta) \le \epsilon \Big\} 
    \nonumber \\
    &\le
    \Prob\Big\{ \Rs_{01}(C_{\tau^*}) > U_{\text{Binom}}(k_{\tau^*}; n, \delta) \Big\}
    \le \delta,
    \label{thm:bg:pps:point}
\end{align}
where 
the last equality in (\ref{thm:bg:pps:mono}) holds
as $\mathbbm{1}\( \y \notin C_\tau(\x)\)$ and $U_{\text{B}}$ are non-decreasing in $\tau$ (\ie Lemma 2 in \cite{Park2022PAC})
and
the last inequality in (\ref{thm:bg:pps:point}) is due to the property of the binomial tail bound $U_{\text{Binom}}$.

\section{Proof of \autoref{lem:fdrbound}}
\label{apdx:proof:lem:fdrbound} 

Since (E) in (\ref{eq:fdrdecomp-ss-ssl}) is decomposed into three terms in \autoref{lem:fdrbounddecomp}, we first find upper bounds on each of the terms and take the union bound as follows. This will return a single upper bound on (E) in (\ref{eq:fdrdecomp-ss-ssl}), which we denote $U_{\text{SSL}}$.

\para{FER Bound.}
First, recall that 
\begin{align*}
    \mathcal{R}_{\text{FER}}(\hat{E}) 
    \coloneqq 
    \mathbbm{P}_{\mathcal{D}_{\hat{S}}}\{ e=0 \wedge G(\x) \in \hat{E}(\y) \}.
\end{align*}

Learning $\hat{E}$ via $\As_\texttt{FER}$ is equivalent to the PAC prediction set learning algorithm that considers the optimization problem in (\ref{eq:pacpsalg}), where the indicator loss is 
$
\ell_{01}(\hat{E}, \x, \y, e) 
\coloneqq 
\mathbbm{1}( e=0 \wedge G(\x) \in \hat{E}(\y) )
$
and the target model is the entailment scoring function $f_E$. Therefore, by \autoref{thm:pacps}, 
for any $n_E \coloneqq |\Z_E|$, we have
\begin{align}
    \mathbb{P}_{\Z_E} \left\{ \Rs_{\text{FER}}(\Eh) \leq \epsilon_E \right\} 
    &= \sum_{m =1 }^{n_E} \mathbb{P}_{\Z_E} \left\{ \mathcal{R}_{\text{FER}}(\Eh) \le \epsilon_E \;\middle|\; 
    | \hat\Z_E | = m \right\} \cdot 
    \mathbb{P}_{\Z_E} \left\{ | \hat\Z_E | = m \right\} 
    \nonumber \\
    &\geq \sum_{m =1 }^{n_E} (1 - \delta'_E/2) \cdot \mathbb{P}_{\Z_E} \left\{ | \hat\Z_E | = m \right\}
    \label{eq:lem2:proof:second0} \\
    &= 1 - \delta'_E/2.
    \label{eq:lem2:proof:second1}
\end{align}
Note that (\ref{eq:lem2:proof:second0}) holds as the PAC guarantee for conformal prediction holds for any number of samples.

The same bound holds with respect to $\Z$. Specifically, letting 
$
\ell_{\text{FER}}(\Z_E, \Z_U) 
\coloneqq
\mathbbm{1}( \mathcal{R}_{\text{FER}}(\hat{E}) \leq \epsilon_E ),
$
we have
\begin{align}
    \mathbb{P}_\Z \left\{ \mathcal{R}_{\text{FER}}(\Eh) \leq \epsilon_E \right\}
    &= \int \ell_{\text{FER}}(\Z_E, \Z_U) ~\mathrm{d}\mathbb{P}(\Z) \nonumber \\
    &= \int \ell_{\text{FER}}(\Z_E, \Z_U) ~\mathrm{d}\mathbb{P}(\Z_E) \mathrm{d}\mathbb{P}(\Z_U) \nonumber \\
    &\geq \int (1 - \delta_E'/2 ) d\mathbb{P}(\Z_U) \nonumber \\
    &= 1 - \delta_E'/2,
    \label{eq:fdrsecondterm}
\end{align}
where the second equality holds due to the i.i.d. assumption on the calibration data and the inequality holds due to (\ref{eq:lem2:proof:second1}). 
%\SP{is it correct? remove? Note the randomness of $\Z_U$ on $\ell_{\text{FER}}( \Z_E, \Z_U )$ is implicitly induced by the selective generator optimized among the calibration data-dependent hypothesis space.}

\para{FNER Bound.}
Recall
\begin{align*}
    \mathcal{R}_{\text{FNER}}(\hat{E})
    &\coloneqq 
    \mathbbm{P}_{\Ds_{\hat{S}}}\{ e=1 \wedge \hat{e}=0 \}.
\end{align*}
Since our goal is to upper-bound $-\mathcal{R}_{\text{FNER}}(\hat{E})$, we consider a lower bound $\mathcal{R}_{\text{FNER}}(\hat{E})$ as follows  for any $n_E \coloneqq |\Z_E|$:
\begin{align}
    \mathbbm{P}_{\Z_E} & \left\{ 
        \mathcal{R}_{\text{FNER}}(\hat{E}) 
        \geq 
        L_{\text{binom}}( \hat{k}; \vert \hat{\Z}_E \vert, \delta_E'/2 ) 
        \right\} \nonumber \\
        &= \sum_{m=1}^{ n_E } \mathbbm{P}_{\Z_E} \left\{
            \mathcal{R}_{\text{FNER}}(\hat{E}) 
            \geq 
            L_{\text{binom}}( \hat{k}; \vert \hat{\Z}_E \vert, \delta_E'/2 ) 
            \;\middle|\; | \hat{\Z}_E | = m
            \right\} \cdot \mathbbm{P}_{\Z_E} \{ | \hat{\Z}_E | = m \} \nonumber \\
            &\geq \sum_{m=1}^{ n_E } ( 1 - \delta_E'/2 ) \cdot \mathbbm{P}_{\Z_E}\{ | \hat{\Z}_E | =m \}, \nonumber \\
            &= 1- \delta_E'/2 \label{eq:lem2:proof:first1}
\end{align}
where the inequality holds due to the binomial tail bound. The same bound holds when the probability is taken over $\Z$. First, let 
\begin{align*}
    \ell_{\text{FNER}}( \Z_E, \Z_U )
    \coloneqq
    \mathbbm{1} \Big( 
        \mathcal{R}_{\text{FNER}}(\hat{E}) 
        \geq 
        L_{\text{Binom}}( \hat{k}; | \hat{\Z}_E |, \delta_E'/2 ) 
        \Big).
\end{align*}
Then,
\begin{align}
    \mathbbm{P}_{\Z}\{ 
        \mathcal{R}_{\text{FNER}}(\Eh)
        \geq
        L_{\text{Binom}}( \hat{k}; | \hat{\Z}_E |, \delta_E'/2 )        
        \} &= \int \ell_{\text{FNER}}( \Z_E, \Z_U ) d\mathbbm{P}(\Z) \nonumber \\
        &= \int \ell_{\text{FNER}}( \Z_E, \Z_U ) d\mathbbm{P}(\Z_E) d\mathbbm{P}(\Z_U) \nonumber \\
        &\geq \int ( 1 - \delta_E'/2 ) d\mathbbm{P}(\Z_U) \nonumber \\
        &= 1 - \delta_E'/2, \label{eq:fdrfirstterm}
\end{align}
where 
the second equality holds due to the i.i.d. assumption 
and
the inequality holds due to (\ref{eq:lem2:proof:first1}).

\para{NER Bound.} Recall 
\begin{align*}
    \mathcal{R}_{\text{NER}}(\hat{E})
    \coloneqq
    \mathbbm{P}_{\Ds_{\hat{S}}}\{ \hat{e} = 0 \}
    = \mathbbm{P}_{\Ds_{\hat{S}}}\{ G(\x) \notin \hat{E}(\y) \}.
\end{align*}
Then, we upper bound $\mathcal{R}_{\text{NER}}(\hat{E})$ as follows for any $n_U \coloneqq |\Z_U|$:
\begin{align}
    \mathbbm{P}_{\Z_U} & \left\{ 
        \mathcal{R}_{\text{NER}}(\hat{E}) 
        \leq 
        U_{\text{Binom}}(\hat{l}; | \hat{\Z}_U |, \delta_S')
        \right\} \nonumber \\
        &= \sum_{m=1}^{n_U} \mathbbm{P}_{\Z_U} \left\{ 
            \mathcal{R}_{\text{NER}}(\hat{E}) 
            \leq 
            U_{\text{Binom}}( \hat{l}; | \hat{\Z}_U |, \delta_S' )
            \;\middle|\; | \hat{\Z}_U | = m
            \right\} \cdot \mathbbm{P}_{\Z_U}\{ | \hat{\Z}_U | = m \} \nonumber \\
            &\geq \sum_{m=1}^{n_U} ( 1 - \delta_S' ) \cdot \mathbbm{P}_{\Z_U}\{ | \hat{\Z}_U | = m \} \nonumber \\
            &= 1 - \delta_S', \label{eq:lem2:proof:third1}
\end{align}
where the inequality holds due to the binomial tail bound. Again, the same bound holds when the probability is taken over $\Z$. First, let 
\begin{align*}
    \ell_{\text{NER}}( \Z_E, \Z_U )
    \coloneqq
    \mathbbm{1}\Big( 
        \mathcal{R}_{\text{NER}}(\hat{E}) 
        \leq  
        U_{\text{Binom}}( \hat{l}; | \hat{\Z}_U |, \delta_S' )
        \Big)
\end{align*}
Then,
\begin{align}
    \mathbbm{P}_{\Z}\big\{ 
        \mathcal{R}_{\text{NER}}(\hat{E})
        \leq
        U_{\text{Binom}}( \hat{l}; | \hat{\Z}_U |, \delta_S' )
        \big\} &= \int \ell_{\text{NER}}( \Z_E, \Z_U ) d\mathbbm{P}(\Z) \nonumber \\
        &= \int \ell_{\text{NER}}( \Z_E, \Z_U ) d\mathbbm{P}(\Z_U) d\mathbbm{P}(\Z_E) \nonumber \\
        &\geq \int ( 1 - \delta_S' ) d\mathbbm{P}(\Z_E) \nonumber \\
        &= 1 - \delta_S',
        \label{eq:fdrthirdterm}
\end{align}
where the inequality holds due to (\ref{eq:lem2:proof:third1}).

Finally, taking the union bound of (\ref{eq:fdrsecondterm}), (\ref{eq:fdrfirstterm}), and (\ref{eq:fdrthirdterm}) completes the proof.

\section{Proof of \autoref{lemma:fdr-e-bound-optimized}}
\label{sec:proof:lemma:fdr-e-bound-optimized}

Let $U_{\text{SSL}}^{(i)}$ be $U_{\text{SSL}}$ for the $i$-th candidate of $\epsilon_E$ in \autoref{alg:opt_u_ssl}. 
Due to \autoref{lem:fdrbound}, the following holds:
\begin{align*}
    \mathbbm{P}_{\Z} \big\{  
        \mathbbm{P}_{\Ds_{\hat{S}}} \{ e = 0 \} 
        > U_{\text{SSL}}^{(i)})
        \big\} 
        \le (\delta_E'+ \delta_S')/Q.
\end{align*}
% \noindent By taking the union bound,
% \begin{align*}
%     \mathbbm{P}_{\Z} \big\{
%         \exists ~ i \in \{1, \dots, Q\},
%         \mathbbm{P}_{\Ds_{\hat{S}}} \{ e = 0 \} > U_{\text{SSL}}^{(i)}
%         \big\} \le \delta_E' + \delta_S'
%     \end{align*}
\noindent Since $U_{\text{SSL}}^{\text{OPT}} = \underset{i \in [Q]}{\text{min}} ~ U_{\text{SSL}}^{(i)}$, we have
\begin{align*}
    \mathbbm{P}_{\Z} \big\{ 
        \mathbbm{P}_{\Ds_{\hat{S}}} \{ e=0 \} > U_{\text{SSL}}^{\text{OPT}}
    \big\}
    &\le
    \mathbbm{P}_{\Z} \big\{
        \exists ~ i \in \{1, \dots, Q\},
        \mathbbm{P}_{\Ds_{\hat{S}}} \{ e = 0 \} > U_{\text{SSL}}^{(i)}
    \big\}
    \\
    &\le
    \sum_{i=1}^Q
    \mathbbm{P}_{\Z} \big\{
        \mathbbm{P}_{\Ds_{\hat{S}}} \{ e = 0 \} > U_{\text{SSL}}^{(i)}
    \big\}
    \\
    &\le
    \delta_E'+ \delta_S',
\end{align*}
where the second inequality is due to a union bound. 
This completes the proof.

\section{Proof of \autoref{thm:guarantee_additional_delta}}
\label{sec:proof:thm:guarantee}

Let
% $
% \mathcal{H} \coloneqq \{ S_1, \dots, S_{| \mathcal{H} |} \}
% $
$
\mathcal{H}
$
be the calibration set-dependent hypothesis space of selective generators, where $n_{\mathcal{H}} \coloneqq |\Hs|$ is always calibration set independent.
Letting $U^{(i)}$ be the FDR-E bound computed given the $i$-th selective generator $S_i$ in $\mathcal{H}$, we first describe how to derive an upper bound of the FDR-E for a given hypothesis $S_i$.

Since an upper bound of (E) in (\ref{eq:fdrdecomp-ss-ssl}) is proved in \autoref{lemma:fdr-e-bound-optimized}, the remaining parts are (i) to derive upper bounds on the others and (ii) to take the union bound. For proportions of the visibility of textual entailment labels, \ie (B) and (D) in (\ref{eq:fdrdecomp-ss-ssl}), and the FDR-E for the supervised case only using entailment-labeled examples, \ie (C) in (\ref{eq:fdrdecomp-ss-ssl}), the followings hold due to the binomial tail bound:
\begin{align*}
    & \mathbbm{P}_{\Z} \Big\{
        \mathbbm{P}_{\mathcal{D}_{S_i}} \{ v=1 \}
        \leq 
        \underbrace{U_{\text{Binom}} \big( 
            | \hat{\Z}_E | ; | \hat{\Z}_E | + | \hat{\Z}_U |, \delta_W/(2 \times | \mathcal{H} |) 
        \big)
        }_{\coloneqq w_\text{SL}^{(i)}}
    \Big\} \geq 1 - \delta_W/(2 \times | \mathcal{H} |); \\
    & \mathbbm{P}_{\Z} \Big\{
        \mathbbm{P}_{\mathcal{D}_{S_i}} \{ v=0 \}
        \leq 
        \underbrace{
        U_{\text{Binom}} \big( 
            | \hat{\Z}_U | ; | \hat{\Z}_E | + | \hat{\Z}_U |, \delta_W/(2 \times | \mathcal{H} |) 
        \big)
        }_{\coloneqq w_\text{SSL}^{(i)}}
    \Big\} \geq 1 - \delta_W/(2 \times | \mathcal{H} |); \\
    & \mathbbm{P}_{\Z} \Big\{
        \mathbbm{P}_{\mathcal{D}_{S_i}} \{ e=0 \}
        \leq 
        \underbrace{
        U_{\text{Binom}} \big( 
            | \hat{\Z}_E^{e=0} | ; | \hat{\Z}_E |, \delta_S/(2 \times | \mathcal{H} |) 
        \big)
        }_{\coloneqq U_\text{SL}^{(i)}}
    \Big\} \geq 1 - \delta_S/(2 \times | \mathcal{H} |),
\end{align*}
where $\hat{\Z}_E~\text{and}~\hat{\Z}_U$ are defined same as \autoref{lem:fdrbound} does, and 
$
\hat{\Z}_E^{e=0} 
\coloneqq 
\{ (\x, \y, e) \in \hat{\Z}_E \mid e=0 \}.
$
Note that the binomial tail bound is applied to filtered sets by the given selective generator (\eg $\hat{\Z}_E$), but we can use the same bound for the non-filtered set $\Z$, by using the same marginalization technique over the size of a filtered set, as in, \eg (\ref{eq:fdrsecondterm}).

Thus, by taking the union bound along with \autoref{lemma:fdr-e-bound-optimized} when $\delta'_E = \delta_E$ and $\delta'_S = \delta_S/2$,
\begin{align}
    \mathbbm{P}_{\Z} \big\{
        \mathcal{R}_E(S_i) \leq U^{(i)}
    \big\} \geq 1 - (\delta_E + \delta_S + \delta_W)/ | \mathcal{H} |, \label{eq:fdrfinal_decomp2}
\end{align}
where $U_i \coloneqq w_{\text{SL}}^{(i)} U_{\text{SL}}^{(i)} + w_{\text{SSL}}^{(i)} U_{\text{SSL}}^{\text{OPT}^{(i)}}$ is the computed FDR-E bound a given selective generator $S_i$. 
Here, $U_{\text{SSL}}^{\text{OPT}^{(i)}}$ refers to the smallest FDR-E bound of (E) in (\ref{eq:fdrdecomp-ss-ssl}) given the $i$-th selective generator. 

\noindent Since (\ref{eq:fdrfinal_decomp2}) holds for all $S_i \in \mathcal{H}$, and the final bound $\hat{U}$ is chosen among them, this completes the proof by taking an union bound, \ie

\begin{align*}
\Prob_\Z \left\{ \Rs_{E}(\Sh) > \Uh \right\}
&\le
\Prob_\Z \left\{ \exists S_i \in \Hs, \Rs_{E}(S_i) > U_i \right\}
\\
&=
\sum_{k=1}^{n_{\mathcal{H}}}d
\Prob_\Z \left\{ \exists S_i \in \Hs, \Rs_{E}(S_i) > U_i, |\Hs| = k \right\}
\\
&=
\sum_{k=1}^{n_{\mathcal{H}}}
\Prob_\Z \left\{ \exists S_i \in \Hs, \Rs_{E}(S_i) > U_i \mid |\Hs| = k \right\} \Prob_\Z \left\{ |\Hs| = k \right\}
\\
&\le
\sum_{k=1}^{n_{\mathcal{H}}}
\sum_{i=1}^k \Prob_\Z \left\{ \Rs_{E}(S_i) > U_i \mid |\Hs| = k \right\} \Prob_\Z \left\{ |\Hs| = k \right\}
\\
&\le
\sum_{k=1}^{n_{\mathcal{H}}}
\sum_{i=1}^k \(\frac{\delta_E + \delta_S + \delta_W}{k}\) \Prob_\Z \left\{ |\Hs| = k \right\}
\\
&= \delta_E + \delta_S + \delta_W.
\end{align*}

\section{Proof of \autoref{lem:perfectcalent}}
\label{sec:proof:lem:perfectcalent}

% We say that $\Eh$ is perfectly separable with respect to $E_\text{true}$ if
% \begin{align} \label{eq:perfectcaleh}
%     \forall (\x, \y),~G(\x) \in \hat{E}(\y) \Longleftrightarrow G(\x) \in E_{\text{true}}(\y).
% \end{align}
We say $f_M$ is perfectly calibrated with respect to $\Ds$, $G$, $E_\text{true}$ if 
\begin{equation} \label{eq:perfectcallan}
    \mathbb{P}_{\mathcal{D}}\{ G(\x) \in E_\text{true}(\y) ~\vert~ f_M( \x, G(\x)) = t \} ) = t, \forall t.              
\end{equation}
The true discovery rate with respect to $E_\text{true}$  conditioned on $f_M( \x, G(\x) ) \geq \tau_S$, \ie $1 - \text{FDR-E}$, is as follows:
\begin{align}
    \mathbb{P}&\{ G(\x)  \in E_{\text{true}}(\y) ~\vert~ f_M( \x, G(\x) ) \geq \tau_S \} 
    \nonumber\\
    &= \frac{\int_{\tau_S}^1 \mathbb{P}\{ G(\x) \in E_{\text{true}}(\y) ~\vert~ f_M( \x, G(\x) )=t \} \mathbb{P}\{ f_M( \x, G(\x) ) =t \} dt}{\int_{\tau_S}^1 \mathbb{P} \{ f_M( \x, G(\x) ) = t \} dt } \nonumber \\
    &= \frac{\int_{\tau_S}^1 t \mathbb{P} \{ f_M( \x, G(\x) ) = t \} dt }{\int_{\tau_S}^1 \mathbb{P} \{ f_M( \x, G(\x) ) = t \} dt }, \label{eq:callemma:perfect}
\end{align}
where
% (\ref{eq:callemma:perfecteh}) holds as $\Eh$ is perfectly separable, \ie (\ref{eq:perfectcalla}),
and
(\ref{eq:callemma:perfect}) holds as $f_M$ is perfectly calibrated, \ie (\ref{eq:perfectcallan}).

Letting 
$h(t) \coloneqq \mathbb{P} \{ f_M( \x, G(\x) ) = t \}$, 
$H(t) \coloneqq \int_{t}^1 h(t') dt'$, 
$i(t) \coloneqq t \mathbb{P} \{ f_M( \x, G(\x) ) = t \}$, 
and 
$I(t) \coloneqq \int_t^1 i(t') dt'$, since we have $\tau_S \leq \frac{\int_{\tau_S}^1 t \mathbb{P} \{ f_M( \x, G(\x) ) = t \} dt }{\int_{\tau_S}^1 \mathbb{P} \{ f_M( \x, G(\x) ) = t \} dt } \leq 1$, the following holds:
\begin{equation*}
    I(1) - I(\tau_S) \geq \tau_S ( H(1) - H(\tau_S) ).
\end{equation*}
Therefore, 
\begin{align*}
    \frac{d}{d\tau_S} \mathbb{P} \{ G(\x) \in E_{\text{true}}(\y) ~\vert~ f_M( \x, G(\x) ) \geq \tau_S \} &= \frac{d}{d\tau_S} \Bigg\{ \frac{I(1) - I(\tau_S)}{H(1) - H(\tau_S)} \Bigg\} \nonumber \\
    &= \frac{-h(\tau_S) \Big[ \tau_S ( H(1) - H(\tau_S) ) - ( I(1) - I(\tau_S) ) \Big]}{( H(1) - H(\tau_S) )^2} \nonumber \\
    &\geq 0.
\end{align*}

This completes the proof.

Note that the classification problem can be reduced from the special case, \ie $E_{\text{true}}(y) \coloneqq E_{\text{EM}}(y)$, where $\mathcal{Y} \coloneqq \mathcal{W}$ and $E_{\text{EM}}(y) \coloneqq \{ y \} = \arg\max_{w \in \mathcal{W}}~\mathbb{P}(Y=w ~\vert~ \X=\x)$.

\newpage

\section{Additional Experiments}
\label{sec:apdx:addexps}

% \begin{figure}[ht]
%     \subfigure[Alpaca7B]{
%         \includegraphics[width=0.48\linewidth]{figs/alpaca7B_plots.pdf}
%         \label{fig:ueplot-alpaca}
%     }
%     \subfigure[GPT-3.5-turbo]{
%         \includegraphics[width=0.48\linewidth]{figs/gpt3.5_plots.pdf}
%         \label{fig:ueplot-gpt3.5}
%     }
%     \caption{Efficiency and FDR bound plot. The FDR bounds for each selective generator $\Sh$ are computed along with its corresponding selection efficiency. The propose algorithm \texttt{NSEGen} automatically select better scoring functions (between $f_{M_1}$ and $f_{M_2}$) for selective generate to minimize the FDR. The baselines in this figure uses a oracle scoring function (\ie manually-selected scoring function to have a better FDR). In this case, baselines is slightly better, but our method can automatically learn to choose a better scoring function from data.}
% \end{figure}

\begin{figure*}
  \subfigure[supervised methods]{
    \includegraphics[width=0.48\linewidth]{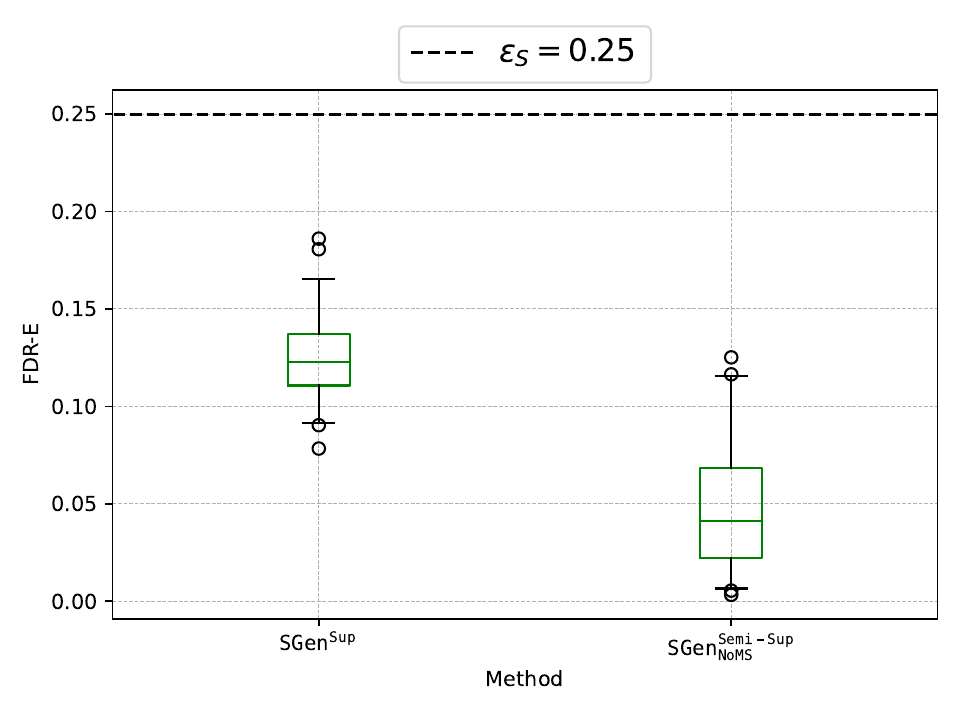}
    \label{fig:gpt-3.5:boxplot:f_M1:SL}
  }
  \subfigure[unsupervised and semi-supervised methods]{
    \includegraphics[width=0.48\linewidth]{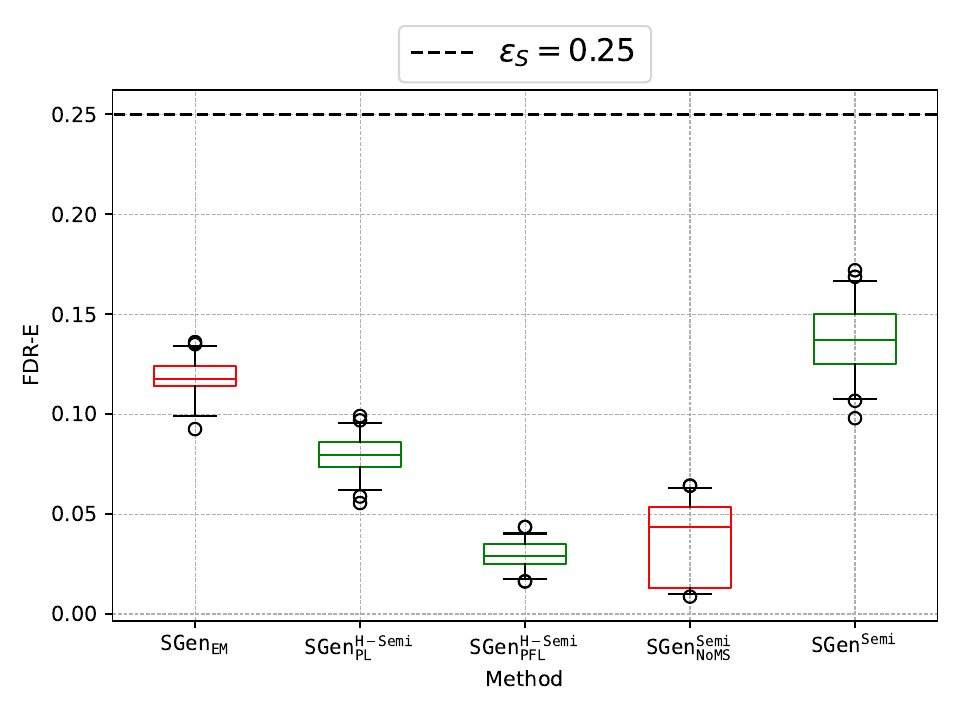}
    \label{fig:gpt-3.5:boxplot:f_M1:SSL}
  }
  % \vspace{-2ex}
  \caption{FDR-E box plots of methods for GPT-3.5-turbo. We randomly split the calibration ad test set 100 times for box plots. For supervised methods (a), we use all entailment labels, \ie $|\Z_E|=|\Z_E^{\text{cal}}|$. For (b), which includes an unsupervised method (\sgem) and semi-supervised methods, we use $|\Z_E|=0.75|\Z_E^{\text{cal}}|$. 
  All methods except for \sgsemi use $f_{M_1}$ as a score function.
  The methods that do not control $\ep_S$ FDR-E in learning at least once are drawn using red boxes but otherwise using green boxes in \autoref{fig:gpt-3.5:boxplot:f_M1:SL} and \autoref{fig:gpt-3.5:boxplot:f_M1:SSL}. 
  We draw the whisker plot to indicate $100\delta \%$ and $100(1-\delta)\%$ quantiles. In both (a) and (b) with green boxes, as the top of the whisker is below of the dotted line, we can see that the FDR-E is well controlled with probability at least $\delta$, \ie they satisfy the PAC guarantee. The numbers of iterations that satisfy $\ep_S$ FDR-E in learning while running 100 iterations are (a) \sgem $= 0$, \sgsup $= 100$, \sgsemisup $= 100$ and (b) \sgpl $= 100$, \sgpfl $= 100$, \sgseminoms $= 18$, \sgsemi $= 100$.
  } 
  
  \label{fig:boxplot:gpt3.5}
\end{figure*}

%% supervised learning results
\begin{table}[ht!]
% \small
\centering

\def\arraystretch{0.9}
\setlength{\tabcolsep}{4pt}

\caption{Comparison results of fully supervised methods. Here, 
we use all entailment labels, \ie 
$|\Z_E| = |\Z_E^{\text{cal}}|$ for GPT-3.5-turbo and Alpaca-7B.
The best results are highlighted in bold, 
results from methods that do not satisfy desired FDR-E guarantee are \underline{underlined}. 
% and
% results where an FDR-E exceeds a desired $\ep$ are highlighted in {\color{red}red}.
In GPT-3.5-turbo and Alpaca-7B, the best efficiency values among methods that satisfy a desired FDR-E guarantee are $0.7535$ and $0.2959$, respectively, which serve as the best achievable efficiency results of semi-supervised methods.
}

\label{tab:alpaca7B:fully-supervised}
\begin{tabular}{cc|cc|cc}
\toprule
\multicolumn{2}{c|}{Models} & \multicolumn{2}{c|}{GPT-3.5-turbo} & \multicolumn{2}{c}{Alpaca-7B} 
\\
\midrule
% GPT-3.5-turbo
\multicolumn{2}{c|}{Methods} & \sgsup & \sgsemisup
% Alpaca-7B
& \sgsup & \sgsemisup
\\
\midrule
\multirow{2}{*}{$f_{M_1}$} & FDR-E
% GPT-3.5-turbo
& $0.1697$ & $0.1066$
% Alpaca-7B
& $0.0400$ & $\underline{0.0231}$
\\
& efficiency
% GPT-3.5-turbo
& $0.6474$ & $0.4657$
% Alpaca-7B
& $0.1769$ & $\underline{0.0922}$
\\
\midrule
\multirow{2}{*}{$f_{M_2}$} & FDR-E
% GPT-3.5-turbo
& $0.2209$ & $0.0914$
% Alpaca-7B
& $0.0983$ & $0.0827$
\\
& efficiency
% GPT-3.5-turbo
& $0.8596$ & $0.5408$
% Alpaca-7B
& $0.4149$ & $0.3675$
\\
\midrule
\midrule
% \multicolumn{2}{c|}{average FDR-E}
% % GPT-3.5-turbo
% & $0.0675$ & $0.1$ & $0.12$
% % Alpaca-7B
% & $0.0329$ & $0.042$ & $0.0262$
% \\
\multicolumn{2}{c|}{average efficiency}
% % GPT-3.5-turbo
% & $0.2750$ & $0.9075$ & $0.4575$
% % Alpaca-7B
% & $0.1159$ & $0.2251$ & $0.1610$
% GPT-3.5-turbo
& $0.7535$ & $0.5033$
% Alpaca-7B
& $0.2959$ & $-$
\\
\bottomrule

\end{tabular}
\end{table}

\begin{table}[ht!]
\centering

\caption{Comparison results of semi-supervised methods. Here, 
$|\Z_U|=10K$ for GPT-3.5-turbo and Alpaca-7B.
The best results are highlighted in \textbf{bold}
and 
results from methods that do not satisfy desired FDR-E guarantee are \underline{underlined}.
We used QA2D dataset, filtered with only SQuAD, where human transformed QA sentences exist.
$\ep = 0.15$.
% (With $\ep_E$ search algorithm, $\delta_P$ and FER control.)
}
\label{tab:gpt3.5:qa2d:semi-supervised}

\setlength{\tabcolsep}{4pt}
% \small

\begin{tabular}{cc|ccccc}
\toprule
\multicolumn{2}{c|}{Models} & \multicolumn{5}{c}{GPT-3.5-turbo}
\\
\midrule
% GPT-3.5-turbo
\multicolumn{2}{c|}{\multirow{2}{*}{Methods}} & \multicolumn{2}{c}{Heuristic} & \multicolumn{3}{c}{Certified}
\\
\cmidrule(lr){3-4} \cmidrule(lr){5-7} 
\multicolumn{2}{c|}{} & \sgpl & \sgpfl & \sgem & \sgseminoms & \sgsemi
\\
\midrule
\midrule
\multirow{2}{*}{$f_{M_1}$} & FDR-E
% GPT-3.5-turbo
& $0.0000$ & ${0.0000}$ & $\underline{0.0213}$ & $0.0962$ & $0.0918$
\\
& efficiency 
% GPT-3.5-turbo
& $0.0387$ & ${0.0227}$ & $\underline{0.4775}$ & $0.8608$ & $0.8502$
\\
\midrule
\multirow{2}{*}{$f_{M_2}$} & FDR-E
% GPT-3.5-turbo
& ${0.0053}$ & ${0.0039}$ & $\underline{0.0831}$ & $0.0169$ & $0.0918$
\\
& efficiency 
% GPT-3.5-turbo
& ${0.1300}$ & ${0.1025}$ & $\underline{0.4862}$ & $0.2156$ & $0.8502$
\\
\midrule\midrule
% \multicolumn{2}{c|}{average FDR-E}
% % GPT-3.5-turbo
% & $0.055$ & $0.065$ & $0.085$
% % Alpaca-7B
% & $0.02475$ & $0.0268$ & $0.0244$
% \\
\multicolumn{2}{c|}{average efficiency}
% GPT-3.5-turbo
%& $0.4950$ & $0.7375$ & $0.4825$ & $0.8200$
& $0.0844$ & $0.0626$ & $-$ & $0.5382$ & $0.8502$
\\
\bottomrule

\end{tabular}
\end{table}

%% supervised learning results
\begin{table}[ht!]
% \small
\centering

\def\arraystretch{0.9}
\setlength{\tabcolsep}{10pt}
\caption{Comparison results of fully supervised methods. Here, 
we use all entailment labels, \ie 
$|\Z_E| = |\Z_E^{\text{cal}}|$ for GPT-3.5-turbo and Alpaca-7B.
The best results are highlighted in bold, 
results from methods that do not satisfy desired FDR-E guarantee are \underline{underlined}.
% In GPT-3.5-turbo and Alpaca-7B, the best efficiency values among methods that satisfy a desired FDR-E guarantee are $0.825$ and $0.3349$, respectively, which serve as the best achievable efficiency results of semi-supervised methods.
We used QA2D dataset, filtered with only SQuAD, where human transformed QA sentences exist.
$\ep = 0.15$.
% (With $\ep_E$ search algorithm, $\delta_P$ and FER control.)
}

\label{tab:gpt3.5:qa2d:fully-supervised}
\begin{tabular}{cc|cc}
\toprule
\multicolumn{2}{c|}{Models} & \multicolumn{2}{c}{GPT-3.5-turbo}
\\
\midrule
% GPT-3.5-turbo
\multicolumn{2}{c|}{Methods} & \sgsup & \sgsemisup
\\
\midrule
\multirow{2}{*}{$f_{M_1}$} & FDR-E
% GPT-3.5-turbo
& $0.1116$ & $0.0454$
\\
& efficiency
% GPT-3.5-turbo
& $0.8956$ & $0.6525$
\\
\midrule
\multirow{2}{*}{$f_{M_2}$} & FDR-E
% GPT-3.5-turbo
& $0.0459$ & $0.0082$
\\
& efficiency
% GPT-3.5-turbo
& $0.3185$ & $0.1532$
\\
\midrule
\midrule
\multicolumn{2}{c|}{average efficiency}
% GPT-3.5-turbo
& $0.6071$ & $0.4029$
\\
\bottomrule

\end{tabular}
\end{table}

% \clearpage
% \input{scratch}

% \input{old/intro}
% %\input{old/rel}
% %\input{old/bg}
% %\input{old/prob}
% \input{old/method}
% \input{old/eval}
% \input{old/conc}
% \input{old/apdx}

%\input{apdx_old}
%\input{eval_old}

%%%%%%%%%%%%%%%%%%%%%%%%%%%%%%%%%%%%%%%%%%%%%%%%%%%%%%%%%%%%

\end{document}